\documentclass[10pt,journal,compsoc]{IEEEtran}
\ifCLASSOPTIONcompsoc
  \usepackage[nocompress]{cite}
\else
  \usepackage{cite}
\fi
\ifCLASSINFOpdf
\else
\fi
\usepackage[utf8]{inputenc} % allow utf-8 input
\usepackage[T1]{fontenc}    % use 8-bit T1 fonts

\usepackage{epsfig}
\usepackage{graphicx}
\usepackage{amsmath}
\usepackage{amssymb}
\usepackage{enumitem}

\usepackage[breaklinks,colorlinks]{hyperref}
\usepackage[capitalize]{cleveref}
\crefname{section}{Sec.}{Secs.}
\crefname{table}{Table}{Tables}
\crefname{figure}{Fig.}{Figs.}

\usepackage{xcolor}
\usepackage{tcolorbox}

\usepackage[linesnumbered,algo2e,boxed]{algorithm2e}
\graphicspath{{figs/}}
\usepackage[figuresright]{rotating}
\usepackage{amsmath,amssymb,textcomp,subfigure,multirow,upgreek}
\usepackage{booktabs}
\usepackage{color,mdwlist}
\usepackage{tikz}
\usepackage[edges]{forest}
\definecolor{hidden-draw}{RGB}{20,68,106}
\definecolor{hidden-pink}{RGB}{255,245,247}

\makeatletter
\DeclareRobustCommand\onedot{\futurelet\@let@token\@onedot}
\def\@onedot{\ifx\@let@token.\else.\null\fi\xspace}

\def\eg{\emph{e.g}\onedot}

\makeatother

\usepackage{ragged2e}

\usepackage{review}

\usepackage{graphicx,fontawesome}
\graphicspath{{figs/}}

\usepackage{wrapfig}

\usepackage{enumitem}
\usepackage[misc]{ifsym}
\setitemize{noitemsep,topsep=0pt,parsep=0pt,partopsep=0pt}

\begin{document}
\title{Empowering  Multimodal LLMs with External Tools: A Comprehensive Survey}

\author{Wenbin An$^*$,
        Jiahao Nie$^*$,
        Yaqiang Wu,
        Feng Tian\textsuperscript{\Letter},
        Shijian Lu\textsuperscript{\Letter},
        Qinghua Zheng\textsuperscript{\Letter}
        % ~\IEEEmembership{Senior Member,~IEEE}
        \thanks{Wenbin An is with the School of Automation Science and Engineering, Xi'an Jiaotong University, Xi'an, 710049, China. E-mail: wenbinan@stu.xjtu.edu.cn. 
        Jiahao Nie is with the Interdisciplinary Graduate Programme, Nanyang Technological University, Singapore, 639798, Singapore. E-mail: jiahao007@e.ntu.edu.sg.
        Yaqiang Wu is with Lenovo Research, Lenovo, Beijing, 100085, China. E-mail: wuyqe@lenovo.com.
        Feng Tian and Qinghua Zheng are with the School of Computer Science and Technology, Xi'an Jiaotong University, Xi'an, 710049, China. E-mail: \{fengtian,qhzheng\}@mail.xjtu.edu.cn.
        Shijian Lu is with the College of Computing and Data Science, Nanyang Technological University, Singapore, 639798, Singapore. E-mail: shijian.lu@ntu.edu.sg.}
	\thanks{*: Equal contribution. \Letter: Corresponding authors.}
}


\markboth{Journal of \LaTeX\ Class Files,%~Vol.~14, No.~8, 
~November~2024}%
{Shell \MakeLowercase{\textit{et al.}}: Bare Demo of IEEEtran.cls for Computer Society Journals}



\IEEEtitleabstractindextext{%
%\begin{abstract}
%The abstract goes here.
%\end{abstract}
\justify
\begin{abstract}
By integrating the perception capabilities of multimodal encoders with the generative power of Large Language Models (LLMs), Multimodal Large Language Models (MLLMs), exemplified by GPT-4V, have achieved great success in various multimodal tasks, pointing toward a promising pathway to artificial general intelligence.
Despite this progress, the limited quality of multimodal data, poor performance on many complex downstream tasks, and inadequate evaluation protocols continue to hinder the reliability and broader applicability of MLLMs across diverse domains.
Inspired by the human ability to leverage external tools for enhanced reasoning and problem-solving, augmenting MLLMs with external tools (\eg, APIs, expert models, and knowledge bases) offers a promising strategy to overcome these challenges.
In this paper, we present a comprehensive survey on leveraging external tools to enhance MLLM performance. Our discussion is structured along four key dimensions about external tools: (1) how they can facilitate the acquisition and annotation of high-quality multimodal data; (2) how they can assist in improving MLLM performance on challenging downstream tasks; (3) how they enable comprehensive and accurate evaluation of MLLMs; (4) the current limitations and future directions of tool-augmented MLLMs.
Through this survey, we aim to underscore the transformative potential of external tools in advancing MLLM capabilities, offering a forward-looking perspective on their development and applications.
The project page of this paper is publicly available at~\href{https://github.com/Lackel/Awesome-Tools-for-MLLMs}{Tools}.
\end{abstract}
}

\maketitle

\IEEEdisplaynontitleabstractindextext

% \begin{center} \bfseries EDICS Category: 3-BBND \end{center}\begin{center} \bfseries EDICS Category: 3-BBND \end{center}\begin{center} \bfseries EDICS Category: 3-BBND \end{center}\begin{center} \bfseries EDICS Category: 3-BBND \end{center}\begin{center} \bfseries EDICS Category: 3-BBND \end{center}\begin{center} \bfseries EDICS Category: 3-BBND \end{center}\begin{center} \bfseries EDICS Category: 3-BBND \end{center}\begin{center} \bfseries EDICS Category: 3-BBND \end{center}
% \fi
\IEEEpeerreviewmaketitle



\section{Introduction}
By scaling up the size of model parameters and training data, Large Language Models (LLMs)~\cite{chatgpt,touvron2023llama,vicuna2023,yang2025qwen3,liu2024deepseek} have demonstrated remarkable performance and adaptability across a wide range of natural language processing tasks~\cite{loop,kasneci2023chatgpt,dka}. 
Despite these advancements, traditional LLMs are inherently limited to processing linguistic data, which restricts their applicability in real-world scenarios that involve diverse modalities such as vision and audio.
Recently, the research on Multimodal Large Language Models (MLLMs)~\cite{dai2023instructblip,liu2023visual,zhu2025internvl3,bai2025qwen25vl,wu2024deepseek} has been developed extensively and intensively. The trained MLLMs enhance the capabilities of LLMs by incorporating multimodal encoders that can process and integrate visual, textual, and auditory signals, enabling perception, interpretation, and reasoning over rich multimodal information. As a result, they have achieved state-of-the-art performance across a variety of multimodal understanding and reasoning benchmarks~\cite{mllmsurvey,lin2024schedule}, positioning them as a promising step toward the development of artificial general intelligence.

\begin{figure*}[t]
    \vspace{-2mm}
    \includegraphics[width=\linewidth]{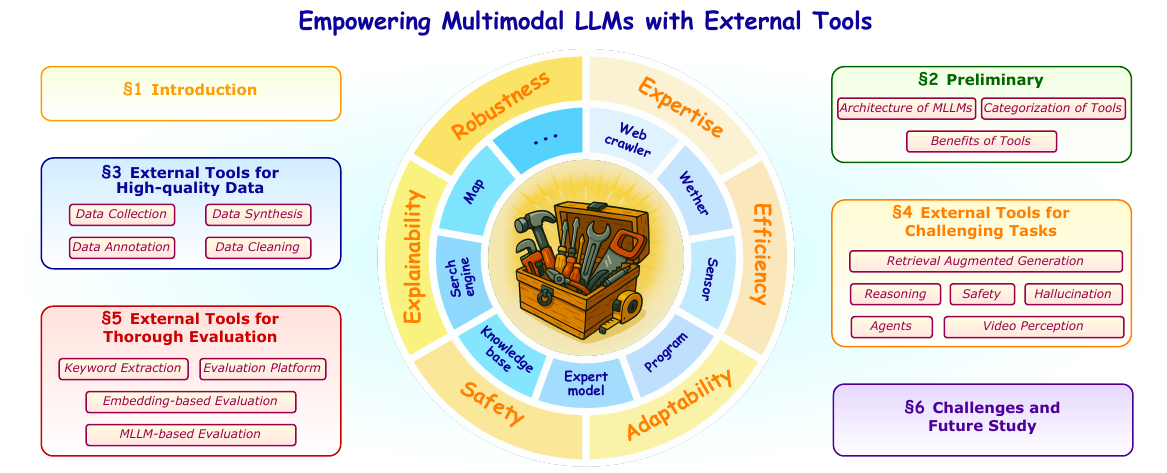}
    \caption{An overview of this survey, which comprises six main components.}
    \label{overview}
    \vspace{-2mm}
\end{figure*}

Despite their great success, significant challenges persist in the development of MLLMs, limiting their reliability and broader applicability across diverse domains. In this paper, we categorize these challenges from three distinctive perspectives. 
The first is the limited quantity and quality of multimodal data. Comprehensive benchmarking is crucial to ensure robust model training and thorough performance evaluation across various tasks. However, unlike textual data, collecting and annotating large-scale, diverse, and high-quality multimodal datasets is challenging due to the complexity and scarcity of well-aligned multimodal data. Such a data bottleneck hinders both effective training and reliable evaluation of MLLMs.
The second is the poor performance over many challenging downstream tasks. Despite recent advancements in simpler tasks such as captioning and basic visual question answering (VQA), MLLMs continue to struggle with complex tasks, including hallucination~\cite{surveyhallucination}, model interpretability~\cite{zhang2024redundancy,sun2024review}, reasoning~\cite{surveyreasoning}, and knowledge-intensive VQA~\cite{dka}. Addressing these challenges is crucial for mitigating risks and ensuring the integrity, reliability, and broader applicability of MLLMs in high-stakes domains such as healthcare~\cite{hu2023advancing} and autonomous driving~\cite{chen2023driving}.
The third is the inadequacy in evaluation protocols. Beyond the aforementioned limited datasets, the widely used evaluation metrics, such as BLEU and CHAIR, often fail to comprehensively capture the quality of MLLM outputs. Given the nature of complex generative multimodal tasks, developing robust metrics that fully assess the answer quality and text-visual consistency remains a key challenge. Such a lack of effective evaluation metrics limits our understanding of model performance, ultimately impeding progress toward more reliable and generalizable MLLMs.
Hence, understanding and addressing the challenges of multimodal data, complicated tasks, and effective evaluation metrics is not only a technical necessity for building more responsible and reliable MLLM systems in real-world applications, but also a crucial step towards achieving artificial general intelligence.

Inspired by the fact that humans use external tools to overcome physical limitations and explore new frontiers, it is natural to leverage the strengths of specialized tools (\eg, domain-specific APIs, expert models, knowledge bases, etc.) to address the aforementioned challenges and advance the development of more capable and reliable MLLMs. For example, using a web crawler in combination with a powerful MLLM such as GPT-4~\cite{gpt4o} can improve the effectiveness and efficiency in acquiring and annotating large-scale and high-quality multimodal data. Similarly, domain-specific knowledge bases and expert models can enhance the performance of MLLMs on challenging downstream tasks. Furthermore, an evaluation platform that incorporates domain-specific models can enable more efficient and comprehensive performance evaluation of MLLMs.
Hence, it has become a crucial task to examine the current progress in MLLM augmentation with external tools to pave the way for future MLLM development. In this paper, we conduct a systematic investigation and comprehensive review of how external tools can help mitigate the issues, aiming to provide a thorough understanding of the key challenges and development directions in this emerging research field.

The following sections of this survey are organized as illustrated in Figure~\ref{overview}. We begin by introducing essential background on MLLM architectures, categorizing external tools, and summarizing the benefits of tool integration (Sec.~\ref{preliminary}). Next, we examine how external tools can be leveraged to obtain high-quality multimodal data, thereby improving MLLM training and evaluation (Sec.~\ref{data}). We then review how external tools enhance MLLM performance on challenging downstream tasks, boosting reliability and applicability across diverse domains (Sec.~\ref{task}). Subsequently, we discuss recent progress in using external tools for MLLM evaluations, which enables more comprehensive and accurate assessment of MLLM performance (Sec.~\ref{evaluation}). Finally, we conclude this survey by summarizing current challenges and highlighting future research directions in this area (Sec.~\ref{future}).
We believe that this survey will deepen our understanding of the limitations of current MLLMs and inspire further research on integrating external tools for building more robust and powerful models.

\section{Preliminary}
\label{preliminary}

In this section, we first introduce the essential components of MLLMs. Next, we categorize the various external tools that can be integrated with MLLMs based on their types and functions. Finally, we summarize the strengths and benefits of tool integration for MLLMs from different perspectives to enhance understanding.

\subsection{Architecture of MLLMs}
A typical MLLM consists of three components: a multimodal encoder, a multimodal projector, and an LLM backbone. The multimodal encoder is responsible for extracting features from the multimodal inputs. The multimodal projector then maps these features into the textual embedding space, ensuring they align with the language model's representations. Finally, the aligned multimodal features are concatenated with the text embeddings and fed into the LLM backbone for response generation~\cite{mllmsurvey}.

\noindent \textbf{Multimodal Encoder} encodes information from multimodal inputs into compact and meaningful representations. By leveraging pre-trained encoders such as CLIP~\cite{clip}, Swin Transformer~\cite{liu2021swin}, and CLAP~\cite{elizalde2023clap}, multimodal encoders effectively extract features from multimodal inputs, thereby facilitating subsequent processing by the LLM backbone.

\begin{figure}[t]
    \centering
    \vspace{-2mm}
    \includegraphics[width=\linewidth]{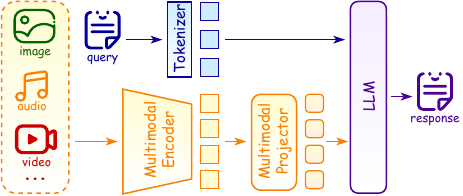}
    \caption{The overall architecture of MLLMs.}
    \vspace{-3mm}
    \label{fig:mllm}
\end{figure}

\noindent \textbf{Multimodal Projector} bridges the gap between textual embeddings and features of other modalities, enabling LLMs to effectively perceive and understand data of different modalities. Most MLLMs employ linear projectors~\cite{liu2023visual}, Q-formers~\cite{dai2023instructblip}, and Perceivers~\cite{laurenccon2024matters} for modality alignment. Through pre-training on large-scale aligned multimodal data, the projector enhances the LLM’s capability to process and integrate multimodal information.

\noindent \textbf{LLM Backbone} is the core component of MLLMs, responsible for processing multimodal data and generating responses. Thanks to large-scale model parameters and expansive training corpora, LLMs are embedded with rich world knowledge and exhibit impressive generalization capabilities. Popular choices include LLaMA~\cite{touvron2023llama}, Qwen~\cite{bai2023qwenllm}, and DeepSeek~\cite{liu2024deepseek}. By combining the perceptual capabilities of multimodal encoders with the generative strengths of LLMs, MLLMs can tackle a broader range of tasks, paving the way for future advancements in multimodal learning.

Combining different components, the core of MLLM is a unified autoregressive framework, formulated as follows:
\begin{equation}
    P(y|x^{mm}, x^{query}) = \prod_{t=1}^T P(y_t \mid y_{<t}, x^{\text{mm}}, x^{\text{query}};\theta)
\end{equation}
where y=$\{y_{1},y_{2},...,y_{t}\}$ represents the sequence of generated tokens, $x^{mm}$ represents the multimodal input, $x^{query}$ represents the input query, and $\theta$ are the model parameters. This formulation enables MLLMs to autoregressively generate outputs conditioned on both the previous tokens and the multimodal context, thereby unifying perception and generation within a single probabilistic modeling framework.

\subsection{Categorization of Tools}
In the LLM domain, the definition of a tool is constrained to a function interface to an external computer program, where LLMs can generate function calls to utilize the tool and thereby enhance their performance~\cite{qin2024tool}. Different from this narrow program-centric definition, we broaden the notion of a tool in this survey to encompass any external means that can enhance the development of MLLMs. In this section, we introduce a taxonomy that categorizes a wide range of tools based on their types and functions as follows:
\begin{itemize}
    \item \textbf{Knowledge Base.} Tools such as knowledge bases empower MLLMs to acquire and integrate external knowledge through retrieval-augmented generation, dynamically expanding their knowledge scope without requiring retraining from scratch.
    \item \textbf{Expert Models.} Powerful LLMs and MLLMs, such as GPT-4V~\cite{gpt4o} and Gemini~\cite{gemini}, can be harnessed for multimodal data synthesis and annotation. Domain-specific expert models, such as grounding models and diffusion models, can enhance MLLM performance on challenging tasks. Moreover, well-aligned expert models such as GPT-4V~\cite{gpt4o} can be leveraged to evaluate the quality of MLLM responses, which augments MLLM performance across multiple aspects.
    \item \textbf{APIs.} Functional APIs (\eg, weather or map APIs) provide real-time updates on weather and geographical information, enabling MLLMs to access timely information for reliable decision-making.
    \item \textbf{Physical Tools.} Physical tools, such as sensors and robots, enable MLLMs to sense and interact with the physical world, thereby expanding their applicability in areas such as autonomous driving, manufacturing, embodied AI, education, and healthcare.
    \item \textbf{Program Tools.} By invoking programming libraries, MLLMs can efficiently complete repetitive tasks. Additionally, employing web crawler programs allows for the automated collection of large-scale multimodal data from the internet, thereby enriching training and evaluation benchmarks.
\end{itemize}

\subsection{Benefits of Tools}
This section presents the multifaceted importance and key benefits of integrating external tools into MLLMs from various perspectives as follows:

\begin{itemize}
\item \textbf{Knowledge Acquisition.} External tools provide access to real-world and domain-specific knowledge, enabling MLLMs to generate more informed and contextually accurate responses across diverse domains.
\item \textbf{Expertise Enhancement.} Incorporating expert models or domain-specific knowledge bases enhances MLLM performance on complex tasks by supplying specialized insights and capabilities.
\item \textbf{Enhanced Efficiency.} Tools such as programmatic libraries and automation frameworks empower MLLMs to perform repetitive or computational tasks more efficiently, improving scalability and productivity.
\item \textbf{Broader Adaptability.} Integration with tools like APIs and sensors allows MLLMs to dynamically adapt to real-time inputs, increasing their applicability across a wider range of real-world scenarios.
\item \textbf{Better Interpretability.} Tools like evaluation frameworks and explanation generators can improve transparency and understanding of MLLM decisions, fostering trust and interpretability.
\item \textbf{Improved Robustness.} External tools can help MLLMs mitigate limitations such as hallucinations or biases by providing reliable references and validation, thus enhancing overall robustness.
\end{itemize}
\section{Data}
\label{data}
\definecolor{hidden-draw}{RGB}{205, 44, 36}
\definecolor{hidden-orange}{RGB}{243,202,120}
\definecolor{hidden-blue}{RGB}{194,232,247}
\definecolor{hidden-yellow}{RGB}{242,244,193}
\definecolor{tree-level-1}{RGB}{245,20,85}
\definecolor{tree-level-2}{RGB}{246,86,118}
\definecolor{tree-level-3}{RGB}{248,177,193}
\definecolor{tree-leaf}{RGB}{176,230,198}
\definecolor{Self}{RGB}{255,0,128}
\definecolor{Ensemble}{RGB}{0,127,255}
\definecolor{Iterative}{RGB}{153,51,255}
\definecolor{exemplar1}{RGB}{136,98,148}
\definecolor{exemplar2}{RGB}{148,210,242}
\definecolor{knowledge1}{RGB}{249,219,152}
\definecolor{knowledge2}{RGB}{255,245,220}

\definecolor{darkblue}{HTML}{000099}
\definecolor{lightblue}{HTML}{D1EEFF}

\definecolor{rootdarkblue}{HTML}{000066}
\definecolor{rootlightblue}{HTML}{C6F3FF}
\definecolor{darkblue}{HTML}{000099}
\definecolor{lightblue}{HTML}{D1EEFF}
\definecolor{darkyellow}{HTML}{FF9B05}
\definecolor{lightyellow}{HTML}{FAFFDD}
\definecolor{darkred}{HTML}{CC0000}
\definecolor{lightred}{HTML}{FFE2DD}
\definecolor{darkgreen}{HTML}{006600}
\definecolor{lightgreen}{HTML}{F3FFE5}
\definecolor{darkorange}{HTML}{FF8000}
\definecolor{lightorange}{HTML}{FFFCD3}
\definecolor{darkpurple}{HTML}{4C0099}
\definecolor{lightpurple}{HTML}{E6D8FF}

\tikzstyle{my-box}=[
    rectangle,
    draw=darkblue,
    rounded corners,
    text opacity=1,
    minimum height=1.5em,
    minimum width=5em,
    inner sep=2pt,
    align=center,
    fill opacity=.2,
]
\tikzstyle{leaf}=[my-box, minimum height=1.5em,
    fill=lightblue, text=black, align=left,font=\scriptsize,
    inner xsep=2pt,
    inner ysep=4pt,
]
\begin{figure*}[!t]
    \centering
    \vspace{-2mm}
    \resizebox{\textwidth}{!}{
        \begin{forest}
            forked edges,
            for tree={
                grow=east,
                reversed=true,
                anchor=base west,
                parent anchor=east,
                child anchor=west,
                base=left,
                font=\sffamily\small,
                rectangle,
                line width=0.8pt,
                draw=darkblue,
                fill=lightblue!60,
                rounded corners,
                align=left,
                minimum width=3em,
                edge+={darkblue, line width=0.8pt},
                s sep=3pt,
                inner xsep=2pt,
                inner ysep=3pt,
                ver/.style={rotate=90, child anchor=north, parent anchor=south, anchor=center},
            },
            where level=1{text width=3.5em,font=\sffamily\scriptsize,edge+={rounded corners}}{},
            where level=2{text width=6.3em,font=\scriptsize,}{},
            where level=3{text width=7.0em,font=\scriptsize,}{},
            where level=4{text width=6.1em,font=\scriptsize,}{},
            [
                \textbf{Multimodal Data}, ver, draw=rootdarkblue,fill=rootlightblue,line width=1.4pt,font=\sffamily,
                    [
                        Data \\ Collection 
                        [
                            VLHTest~\cite{vlhtest}{, }
                            CorrelationQA~\cite{correlationqa}{, }
                            MMVP~\cite{mmvp}{, }
                            Concept~\cite{concept}{, }
                            Flume~\cite{flume}{, }
                            PaLI~\cite{pali}{, }
                            LAION~\cite{laion}{, }
                            DATACOMP~\cite{datacomp}{, }\\
                            ALIGN~\cite{align}{, }
                            MMC4~\cite{mmc4}{, }
                            Wukong~\cite{wukong}{, }
                            WIT~\cite{wit}{, }
                            Conceptual-12M~\cite{conceptual12m}{, }
                            Redcaps~\cite{redcaps}{, }
                            MMArXiv~\cite{mmarxiv}{, }
                            VQA-v2~\cite{vqav2}{, }\\
                            AIC~\cite{aic}{, }
                            Im2Text~\cite{im2text}
                            , leaf, text width=41.5em
                        ]
                    ]
                    [
                        Data \\ Synthesis,draw=darkyellow,fill=lightyellow,edge+={darkyellow, line width=0.8pt},
                        [
                            LongHalQA~\cite{longhalqa}{, }
                            AutoHallusion~\cite{autohallusion}{, }
                            THRONE~\cite{throne}{, }
                            PhD~\cite{phd}{, }
                            AIGCs~\cite{aigcs}{, }
                            VLHTest~\cite{vlhtest}{, }
                            EASY~\cite{easy}{, }
                            MOCHA~\cite{mocha}{, }\\
                            CorrelationQA~\cite{correlationqa}{, }
                            FOHE~\cite{fohe}{, }
                            NOPE~\cite{nope}{, }
                            LRV~\cite{lrv}{, }
                            CIEM~\cite{ciem}{, }
                            MMHalSnowball~\cite{mmhalsnowball}{, }
                            SILKIE~\cite{silkie}{, }
                            MMArXiv~\cite{mmarxiv}{, }\\
                            DVQA~\cite{dvqa}{, }
                            MMRel~\cite{mmrel}{, }
                            LLaVA~\cite{liu2023visual}{, }
                            Invest~\cite{invest}
                            , leaf, text width=41.5em,draw=darkyellow,fill=lightyellow,edge+={darkyellow, line width=0.8pt},
                        ]
                    ]
                                        [
                       Data \\ Annotation,draw=darkred,fill=lightred,edge+={darkred, line width=0.8pt}, 
                        [
                            VGA~\cite{vga}{, }
                            ALOHa~\cite{aloha}{, }
                            PhD~\cite{phd}{, }
                            HalEval~\cite{haleval}{, }
                            EasyDetect~\cite{easydetect}{, }
                            MOCHA~\cite{mocha}{, }
                            MERLIM~\cite{merlim}{, }
                            RAH~\cite{rah}{, }
                            MMRel~\cite{mmrel}{, }\\
                            HallE-Control~\cite{hallecontrol}{, }
                            HAELM~\cite{haelm}{, }
                            POPE~\cite{pope}{, }
                            SoM~\cite{som}{, }
                            M-HalDetect~\cite{mhaldetect}{, }
                            HALLUGEN~\cite{hallucinogen}{, }
                            Im2Text~\cite{im2text}{, }
                            PaLI~\cite{pali}
                            , leaf, text width=41.5em,draw=darkred,fill=lightred,edge+={darkred, line width=0.8pt},
                        ]
                    ]
                    [
                       Data \\ Cleaning,draw=darkgreen,fill=lightgreen,edge+={darkgreen, line width=0.8pt},  
                        [
                            VideoHallucer~\cite{videohallucer}{, }
                            IllusionVQA~\cite{illusionvqa}{, }
                            AIGCs~\cite{aigcs}{, }
                            LRV~\cite{lrv}{, }
                            VDGD~\cite{vdgd}{, }
                            VIDHALLUC~\cite{vidhalluc}{, }
                            Concept~\cite{concept}{, }\\
                            LAION~\cite{laion}{, }
                            DATACOMP~\cite{datacomp}{, }
                            COYO~\cite{coyo}{, }
                            OBELICS~\cite{obelics}{, }
                            MMC4~\cite{mmc4}{, }
                            Wukong~\cite{wukong}{, }
                            WIT~\cite{wit}{, }
                            Conceptual-12M~\cite{conceptual12m}{, }\\
                            DVQA~\cite{dvqa}{, }
                            Redcaps~\cite{redcaps}{, }
                            MMArXiv~\cite{mmarxiv}
                            , leaf, text width=41.5em,draw=darkgreen,fill=lightgreen,edge+={darkgreen, line width=0.8pt},
                        ]
                    ]
            ]
        \end{forest}
    }
    \vspace{-5mm}
    \caption{A summary of external tools for high-quality multimodal data acquisition.}
    \vspace{-3mm}
    \label{taxo_of_data}
\end{figure*}
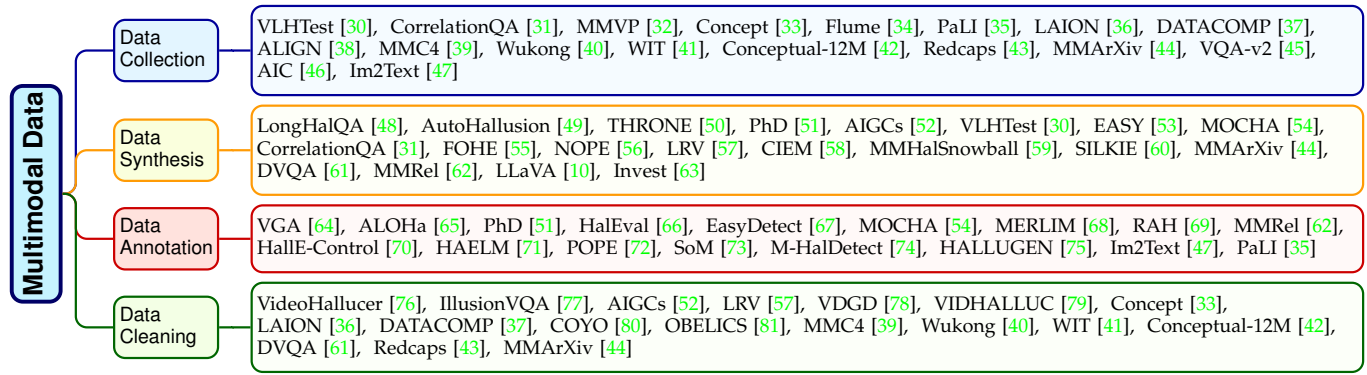

Due to the complexity and scarcity of well-aligned multimodal data, collecting and annotating large-scale, diverse, and high-quality datasets remains both critical and challenging. This limitation poses significant obstacles to the effective training and reliable evaluation of MLLMs.
In this section, we review how external tools can be leveraged to facilitate high-quality multimodal data acquisition from four key perspectives: \textit{Data Collection}, \textit{Data Synthesis}, \textit{Data Annotation}, and \textit{Data Cleaning}, as illustrated in Fig.~\ref{taxo_of_data}.

\subsection{Data Collection}
Employing external tools significantly enhances the effectiveness and scalability of collecting large-scale multimodal data from the Internet. 
Many large-scale datasets designed for MLLM pre-training have leveraged different tool chains and infrastructures to streamline the data collection process. For instance,
Concept~\cite{concept}, ALIGN~\cite{align}, WIT~\cite{wit}, Conceptual-12M~\cite{conceptual12m}, and PaLI~\cite{pali} employ the Flume library~\cite{flume} to process billions of Internet webpages in parallel and collect image-caption pairs from the webpages.
DATACOMP~\cite{datacomp} utilizes cc2dataset, an Apache Spark-based library, to extract pairs of image URLs and nonempty alt-text from all Common Crawl snapshots to collect image-text data pairs.
LAION~\cite{laion} collects multimodal data from the Common Crawl website by downloading the raw images with asynchronous requests using the Trio and Asks Python libraries.
CorrelationQA~\cite{correlationqa} employs a search engine to collect realistic images from the Internet to ensure that the collected data is not biased.
Wukong~\cite{wukong} and AIC~\cite{aic} take a series of keywords as the starting point and employ a search engine to search for images and corresponding captions.

Several studies also employ external tools to collect data for various downstream tasks. For example, VLHTest~\cite{vlhtest} employs CLIP~\cite{clip} and DINO-v2~\cite{oquab2023dinov2} to compute similarities of multimodal data and collect hallucination candidates.
MMVP~\cite{mmvp} collects CLIP-blind data pairs by discovering image pairs that are proximate in CLIP~\cite{clip} feature space but distant in DINO-v2~\cite{oquab2023dinov2} feature space.
MMC4~\cite{mmc4} first collects a set of images for each document, and then associates each image with a sentence by formulating a bipartite assignment problem with CLIP~\cite{clip}.
Redcaps~\cite{redcaps} constructs a manually curated set of subreddits and collects subreddits with a high volume of image posts. 
MMArXiv~\cite{mmarxiv} collects academic papers from ArXiv by extracting images and captions from the source file. Then, a robust tool, ImageMagick, is employed to convert images into JPEG format for easy processing.
VQA-v2~\cite{vqav2} uses a CNN to extract image features and computes similarities between images to find complementary images with high similarity.
Im2Text~\cite{im2text} queries Flickr using a huge number of pairs of terms to collect a large initial set of photographs with associated text.

\subsection{Data Synthesis}
In addition to collecting multimodal data from the Internet, an alternative approach is to synthesize multimodal data using powerful LLMs and MLLMs. 
Most existing work can be broadly grouped into two categories, namely, textual synthesis based on visual inputs and visual synthesis based on textual inputs. 
Specifically, several studies leverage LLMs to generate synthetic textual data grounded in visual content. For instance, 
LongHalQA~\cite{longhalqa} prompts GPT-4~\cite{gpt4o} to generate hallucinatory data with multi-round long conversations, leading to more reliable and efficient evaluations for long context hallucination.
THRONE~\cite{throne} employs an external LLM to extract objects and answers from image captions, which can quantitatively evaluate hallucinations in MLLM free-form outputs.
EASY~\cite{easy} employs GPT-4~\cite{gpt4o} to generate deceptive prompts that would intentionally mislead MLLMs for hallucination evaluation.
FOHE~\cite{fohe} employs ChatGPT~\cite{chatgpt} to extract keywords from the original captions and generate rewritten captions based on the given keywords.
NOPE~\cite{nope} and LRV~\cite{lrv} explore different methods with GPT-4~\cite{gpt4o} to generate negative questions based on the images and captions.
CIEM~\cite{ciem} employs ChatGPT~\cite{chatgpt} to generate questions and chain-of-thought reasons, matching how humans think of and answer questions.
MMHalSnowball~\cite{mmhalsnowball} utilizes ChatGPT~\cite{chatgpt} to allocate hallucination type, generate hallucinatory answers, and the corresponding hallucinatory descriptions.
SILKIE~\cite{silkie} collects instructions from various sources and decodes the corresponding responses using 4 models sampled from the LVLM pool to generate a vision-language feedback dataset.
MMArXiv~\cite{mmarxiv} designs a prompt template to query GPT-4V~\cite{gpt4o} for generating QA pairs based on the images. 
LLaVA~\cite{liu2023visual} leverages ChatGPT~\cite{chatgpt} to create instruction-following data involving visual content.
While Invest~\cite{invest} further prompts ChatGPT~\cite{chatgpt} to manipulate associated captions to create positive and negative training samples.

Conversely, several studies focus on synthesizing visual data from textual inputs. 
For instance, MOCHA~\cite{mocha} uses captions from MS-COCO as seeds to generate diverse synthetic captions with LLMs~\cite{chatgpt}, and then uses Stable Diffusion~\cite{stablediffusion} to generate images corresponding to the captions.
AutoHallusion~\cite{autohallusion}, AIGCs~\cite{aigcs}, VLHTest~\cite{vlhtest}, CorrelationQA~\cite{correlationqa}, MMRel~\cite{mmrel}, and PhD~\cite{phd} employ image generation models, such as DALLE-3~\cite{dalle} and Stable Diffusion~\cite{stablediffusion}, to create images based on the given contextual elements.
In addition to leveraging image generation models, DVQA~\cite{dvqa} uses Python’s popular drawing tool, Matplotlib, to generate charts with unparalleled programmatic control over each of the elements drawn.

\subsection{Data Annotation}
With the rapid growth of visual data, external tools have been widely adopted to annotate images to construct well-aligned multimodal datasets. These tools facilitate scalable and high-quality annotations that support tasks such as hallucination evaluation, object grounding, and reasoning.

Several works utilize large models such as GPT-4V~\cite{gpt4o} to directly annotate content in the visual data. For example, VGA~\cite{vga}, HalEval~\cite{haleval}, RAH~\cite{rah}, HAELM~\cite{haelm}, SoM~\cite{som}, MMRel~\cite{mmrel}, and PhD~\cite{phd} prompt GPT-4\cite{gpt4o} to analyze and annotate image content, resulting in high-quality comprehension datasets tailored for hallucination evaluation in MLLMs.
Object-centric annotation is also prevalent. For example, Invest~\cite{invest} and POPE~\cite{pope} use SEEM~\cite{seem} to detect objects within images. ALOHa~\cite{aloha} combines an LLM with object detection to extract groundable objects from captions, assess their semantic similarity to detected entities, and use Hungarian matching to compute hallucination scores.

To enable more effective and efficient annotation, more complex tool chains are also employed. 
For example, EasyDetect~\cite{easydetect} integrates multiple tools, including Grounding DINO~\cite{groundingdino}, GPT-4~\cite{gpt4o}, Google Search API, and MAERec, to annotate multimodal data for hallucination detection. 
MERLIM~\cite{merlim} prompts the ChatGPT API to select an object category and generate semantically meaningful negative relationships. 
HallE-Control~\cite{hallecontrol} first categorizes MSCOCO ground-truth objects using the open-vocabulary detector RAM~\cite{ram}, and then uses GPT-4V~\cite{gpt4o} to synthesize captions based on grounded object lists.
Several efforts also incorporate advanced vision-language models. For example, M-HalDetect~\cite{mhaldetect} prompts InstructBLIP~\cite{dai2023instructblip} to generate hallucination detection responses, forming a labeled dataset. HALLUGEN~\cite{hallucinogen} targets the medical domain by extracting disease categories from radiology reports using NLP techniques, and then uses ChatGPT~\cite{chatgpt} to annotate corresponding medical images.

In addition to generative models, rule-based or automated services are also widely adopted. For example, PaLI~\cite{pali} leverages a publicly available OCR system to extract annotations from images, generating 29 billion image–OCR pairs. Im2Text~\cite{im2text} adopts a retrieval-based annotation strategy: it computes the global similarity between a query image and a large collection of captioned web images, retrieves the closest match, and transfers the corresponding caption as the annotation.

\subsection{Data Cleaning}
After collecting and annotating multimodal data, employing external tools to clean out low-quality or noisy samples is crucial for ensuring data reliability and annotation integrity. 
Many studies have adopted different filtering strategies using expert models. 
For instance, VideoHallucer~\cite{videohallucer} applies CLIP~\cite{clip} to compute semantic similarity between video segments and selects those with scores above 0.85 for downstream annotation. Similarly, AIGCs~\cite{aigcs} measures the cosine similarity between synthetic images and their corresponding textual descriptions using CLIP~\cite{clip}, discarding samples with low alignment.
IllusionVQA~\cite{illusionvqa} ensures textual safety by employing the Perspective API to filter out toxic content in questions, answers, and choices. 
LRV~\cite{lrv} and VDGD~\cite{vdgd} leverage GPT-4V~\cite{gpt4o} to evaluate dataset quality and remove inconsistent or irrelevant samples. 
VIDHALLUC~\cite{vidhalluc} combines CLIP~\cite{clip} and DINO-v2~\cite{oquab2023dinov2} for semantic and visual similarity filtering, and further utilizes GPT-4V~\cite{gpt4o} to eliminate videos with redundant actions.

Large-scale datasets created for pre-training also incorporate multi-stage cleaning pipelines. For instance, Concept~\cite{concept}, Conceptual-12M~\cite{conceptual12m}, Wukong~\cite{wukong}, WIT~\cite{wit}, and COYO~\cite{coyo} apply a three-stage cleaning strategy, namely image-based, text-based, and joint image-text based, to remove noisy samples. 
LAION~\cite{laion} filters samples using Google’s CLD3 API for language detection and CLIP~\cite{clip} for image-text similarity filtering. 
Similarly, DATACOMP~\cite{datacomp} employs the Detoxify library to exclude unsafe text and CLIP to filter out samples with explicit visual content.

In addition to similarity-based cleaning, many studies further incorporate additional quality checks. For instance, OBELICS~\cite{obelics} uses FastText to remove non-English content and integrates the Spawning API to eliminate copyright-violating images. MMC4~\cite{mmc4} employs an MLP to remove samples with NSFW probabilities exceeding 0.1. 
RedCaps~\cite{redcaps} and MMArXiv~\cite{mmarxiv} implement post-filtering strategies on both images and captions to improve the quality of user-generated content. 
Lastly, DVQA~\cite{dvqa} targets question bias by removing questions with spurious correlations between object styles, colors, and labels.
\begin{figure*}[t]
    \vspace{-2mm}
    \includegraphics[width=\linewidth]{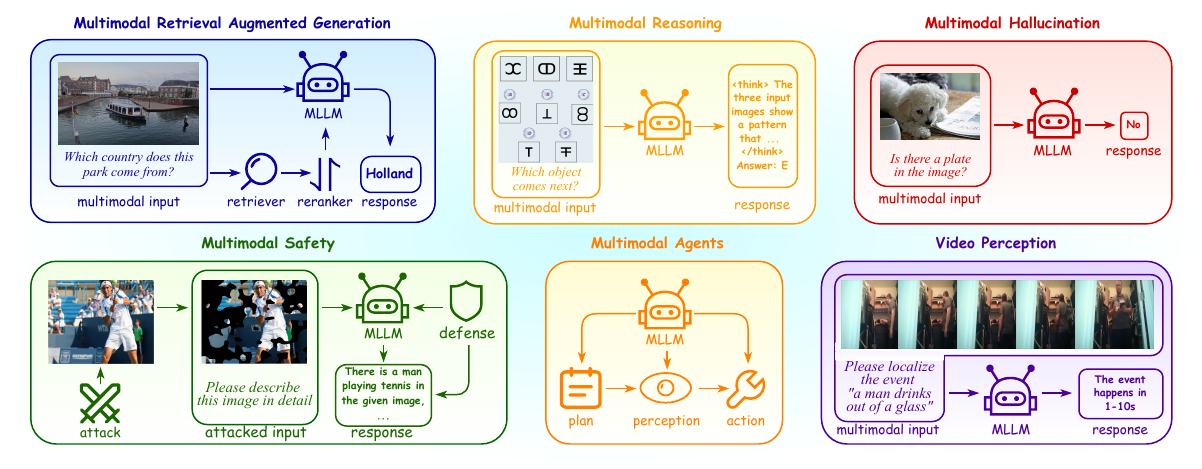}
    \vspace{-6mm}
    \caption{An overview of the six challenging multimodal tasks discussed in this survey.}
    \label{taskoverview}
    \vspace{-2mm}
\end{figure*}

\section{Tasks}
\label{task}

Despite recent progress on relatively simple tasks such as image captioning and basic visual question answering, MLLMs still face significant challenges when dealing with more complex tasks. In this section, we provide a comprehensive review of how external tools have been employed to enhance MLLM performance across six categories of advanced tasks: \textit{Multimodal Retrieval-Augmented Generation}, \textit{Multimodal Reasoning}, \textit{Multimodal Hallucination}, \textit{Multimodal Safety}, \textit{Multimodal Agents}, and \textit{Video Perception}. Fig.~\ref{taskoverview} provides an overview of these six tasks, while Fig.~\ref{taxo_of_model} illustrates a taxonomy of representative studies within each category.

\subsection{Multimodal Retrieval Augmented Generation}
Multimodal Retrieval-Augmented Generation (MRAG) aims to improve the factual accuracy of Multimodal Large Language Models (MLLMs) and reduce hallucinations by retrieving and incorporating knowledge from external knowledge bases. Typically, MRAG systems adopt a three-stage pipeline: (1) a retriever employs embedding models to identify relevant knowledge from external sources; (2) a reranker applies ranking techniques to improve the precision of the retrieved content; and (3) the selected knowledge is integrated into the generation process of MLLMs to produce more grounded and factual responses.
In this section, we review how external tools are utilized to enhance MRAG performance across the three stages.

\subsubsection{Knowledge Retrieval}
MRAG typically incorporates an additional multimodal retriever to encode input queries and retrieve relevant knowledge from external knowledge bases. The multimodal retriever, such as ALIGN~\cite{align}, CLIP~\cite{clip}, BLIP~\cite{li2022blip}, and GME~\cite{gme}, is trained to maximize the similarity between the embeddings of multimodal queries and corresponding knowledge items, thereby enabling more accurate retrieval.

Beyond traditional text-based retrieval strategies~\cite{contriever}, vision-centric approaches leverage image representations for retrieval. 
For example, Pic2word~\cite{pic2word} transforms images into textual tokens to support multimodal retrieval. 
XL-HeadTags~\cite{xlheadtags} incorporates auxiliary cues such as embedded images and captions from articles to improve sentence retrieval. 
Echosight~\cite{echosight} retrieves visually similar content by using reference images directly as queries. 
eCLIP~\cite{eclip} extends CLIP~\cite{clip} with a heatmap processor and mixup augmentation, enhancing retrieval performance in annotation-scarce domains.
VISA~\cite{visa} enhances transparency by highlighting evidence regions in retrieved passages using bounding boxes, improving interpretability. 

In addition to static images, several methods extend MRAG to other modalities such as videos and audio. For instance, VideoRAG~\cite{videorag}, DRVideo~\cite{drvideo}, CTCH~\cite{ctch}, and OMAgent~\cite{omagent} perform retrieval for videos, while ReCAP~\cite{recap}, WavRAG~\cite{wavrag}, SpeechRAG~\cite{speechrag}, and TTA-RAG~\cite{ttarag} support retrieval for audio-based queries.

\subsubsection{Knowledge Reranking}
Although retrieval strategies and multimodal retrievers have seen significant progress, the retrieved knowledge is usually noisy due to the substantial modality gap. To improve the precision of retrieval, various reranking techniques have been proposed to identify and retain truly relevant knowledge from all the retrieved candidates.

For example, MIS~\cite{mis} introduces a Multi-modal Information Similarity (MIS) metric to assess the alignment between input queries and candidate knowledge, enabling effective reranking and noise suppression. 
MSIER~\cite{msier} leverages CIDEr scores to rerank candidates and select positive and negative samples for retriever training. RULE~\cite{rule} incorporates a factuality risk control strategy to dynamically determine the amount of knowledge used during inference, filtering noisy data while minimizing information loss.
RAMM~\cite{ramm} adjusts knowledge sampling probabilities based on similarity scores to balance diversity and quality. 
RS~\cite{rs} introduces a relevancy score from an expert model to select more pertinent knowledge, while RAG-Check~\cite{ragcheck} combines relevancy and correctness scores for comprehensive knowledge quality assessment. 
mRRAG~\cite{mrrag} employs relevance-reflection tokens to score retrieved results, enhancing final answer ranking.

Several approaches integrate Large Language Models (LLMs) into the reranking pipeline. 
For instance, OMG-QA~\cite{omgqa} and LDRE~\cite{ldre} use LLMs~\cite{chatgpt} to evaluate knowledge quality and generate divergent captions, respectively. 
UniRAG~\cite{unirag} first converts multimodal input into text and reranks documents using language models. 
EgoInstructor~\cite{egoinstructor} enhances retrieval by exploiting cross-view video clips, and GME~\cite{gme} filters out low-relevance image-caption pairs using a CLIP model~\cite{clip}.
MM-Embed~\cite{mmembed} incorporates LLaVA to frame reranking as a series of True/False questions, while MAIN~\cite{main} employs an agent-based schema to judge document relevance. 
Finally, EchoSight~\cite{echosight} fine-tunes the Q-Former and text encoder of the retriever to rerank retrieved items more effectively.

\definecolor{hidden-draw}{RGB}{205, 44, 36}
\definecolor{hidden-orange}{RGB}{243,202,120}
\definecolor{hidden-blue}{RGB}{194,232,247}
\definecolor{hidden-yellow}{RGB}{242,244,193}
\definecolor{tree-level-1}{RGB}{246,20,85}
\definecolor{tree-level-2}{RGB}{246,86,118}
\definecolor{tree-level-3}{RGB}{248,177,193}
\definecolor{tree-leaf}{RGB}{176,230,198}
\definecolor{Self}{RGB}{255,0,128}
\definecolor{Ensemble}{RGB}{0,127,255}
\definecolor{Iterative}{RGB}{153,51,255}
\definecolor{exemplar1}{RGB}{136,98,148}
\definecolor{exemplar2}{RGB}{148,210,242}
\definecolor{knowledge1}{RGB}{249,219,152}
\definecolor{knowledge2}{RGB}{255,246,220}

\definecolor{rootdarkblue}{HTML}{000066}
\definecolor{rootlightblue}{HTML}{C6F3FF}
\definecolor{darkblue}{HTML}{000099}
\definecolor{lightblue}{HTML}{D1EEFF}
\definecolor{darkyellow}{HTML}{FF9B05}
\definecolor{lightyellow}{HTML}{FAFFDD}
\definecolor{darkred}{HTML}{CC0000}
\definecolor{lightred}{HTML}{FFE2DD}
\definecolor{darkgreen}{HTML}{006600}
\definecolor{lightgreen}{HTML}{F3FFE5}
\definecolor{darkorange}{HTML}{FF8000}
\definecolor{lightorange}{HTML}{FFFCD3}
\definecolor{darkpurple}{HTML}{4C0099}
\definecolor{lightpurple}{HTML}{E6D8FF}

\tikzstyle{my-box}=[
    rectangle,
    draw=darkblue,
    rounded corners,
    text opacity=1,
    minimum height=1pt,
    minimum width=1pt,
    inner sep=2pt,
    align=center,
    fill opacity=.2,
]
\tikzstyle{leaf}=[my-box, minimum height=1.5em,
    text=black, align=left,font=\scriptsize,
    inner xsep=2pt,
    inner ysep=4pt,
]
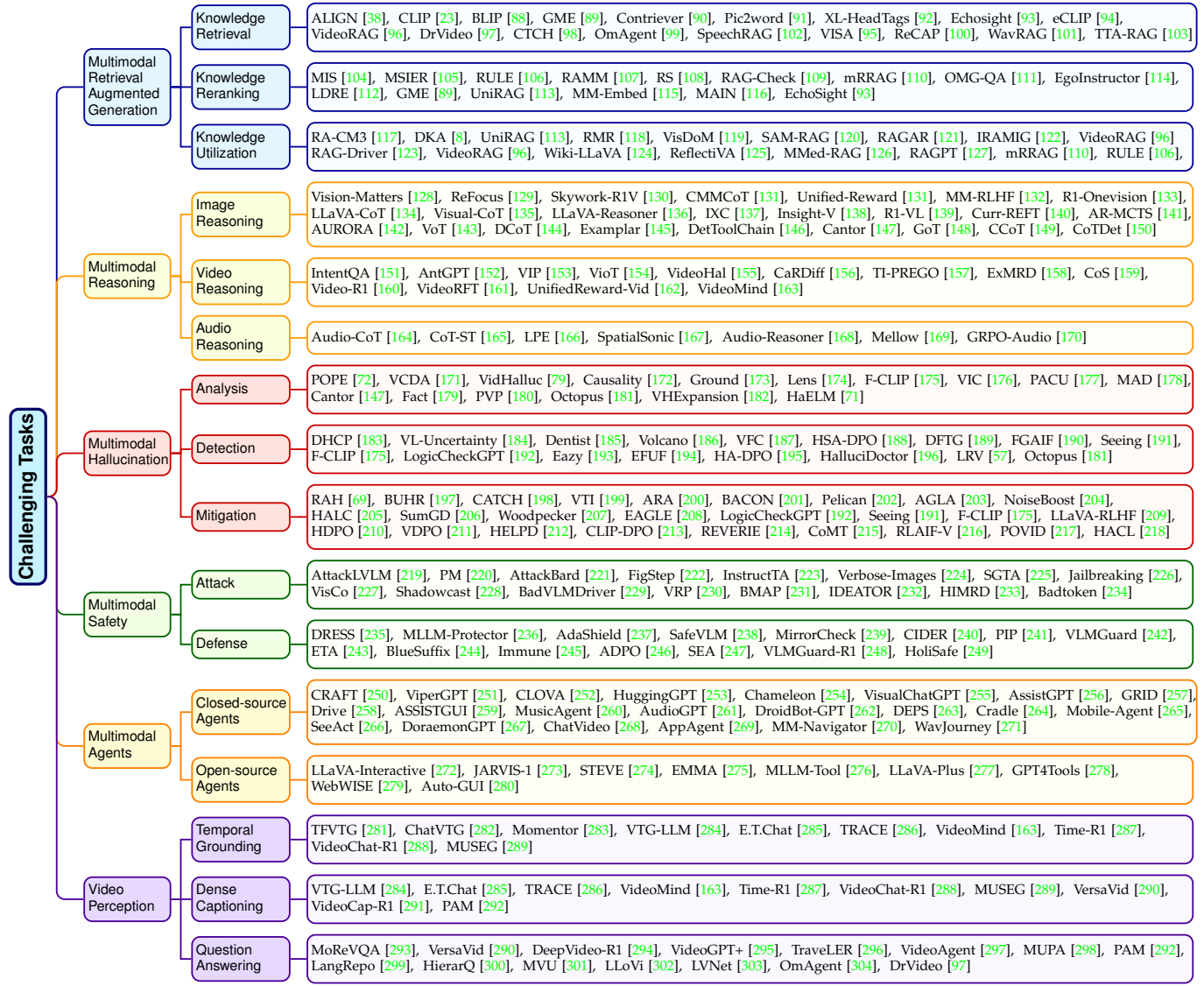
\begin{figure*}[!t]
    \centering
    \resizebox{\textwidth}{!}{
        \begin{forest}
            forked edges,
            for tree={
                grow=east,
                reversed=true,
                anchor=base west,
                parent anchor=east,
                child anchor=west,
                base=left,
                font=\sffamily\small,
                rectangle,
                line width=0.8pt,
                draw=darkblue,
                fill=lightblue!60,
                rounded corners,
                align=left,
                minimum width=2em,
                edge+={darkblue, line width=0.8pt},
                s sep=5pt,
                inner xsep=2pt,
                inner ysep=3pt,
                ver/.style={rotate=90, child anchor=north, parent anchor=south, anchor=center,},
            },
            where level=1{text width=4.1em,font=\sffamily\scriptsize,edge+={rounded corners},l sep=10pt,}{},
            where level=2{text width=4.7em,font=\sffamily\scriptsize,l sep=8pt,}{},
            where level=3{text width=3.6em,font=\scriptsize,}{},
            where level=4{text width=6.1em,font=\scriptsize,}{},
            [
                \textbf{Challenging Tasks}, ver, l sep=1pt, draw=rootdarkblue,fill=rootlightblue,line width=1.4pt,font=\sffamily,
                    [   
                        Multimodal \\Retrieval \\Augmented \\Generation
                        [
                            Knowledge \\Retrieval
                            [
                                ALIGN~\cite{align}{, }
                                CLIP~\cite{clip}{, }
                                BLIP~\cite{li2022blip}{, }
                                GME~\cite{gme}{, }
                                Contriever~\cite{contriever}{, }
                                Pic2word~\cite{pic2word}{, }
                                XL-HeadTags~\cite{xlheadtags}{, }
                                Echosight~\cite{echosight}{, }
                                eCLIP~\cite{eclip}{, }\\
                                VideoRAG~\cite{videorag}{, }
                                DrVideo~\cite{drvideo}{, }
                                CTCH~\cite{ctch}{, }
                                OmAgent~\cite{omagent}{, }
                                SpeechRAG~\cite{speechrag}{, }
                                VISA~\cite{visa}{, }
                                ReCAP~\cite{recap}{, }
                                WavRAG~\cite{wavrag}{, }
                                TTA-RAG~\cite{ttarag}
                                , leaf, text width=46.2em
                            ]
                        ]
                        [
                            Knowledge \\Reranking
                            [
                                MIS~\cite{mis}{, }
                                MSIER~\cite{msier}{, }
                                RULE~\cite{rule}{, }
                                RAMM~\cite{ramm}{, }
                                RS~\cite{rs}{, }
                                RAG-Check~\cite{ragcheck}{, }
                                mRRAG~\cite{mrrag}{, }
                                OMG-QA~\cite{omgqa}{, }
                                EgoInstructor~\cite{egoinstructor}{, }\\
                                LDRE~\cite{ldre}{, }
                                GME~\cite{gme}{, }
                                UniRAG~\cite{unirag}{, }
                                MM-Embed~\cite{mmembed}{, }
                                MAIN~\cite{main}{, }
                                EchoSight~\cite{echosight}
                                , leaf, text width=46em
                            ]
                        ]
                        [
                            Knowledge \\Utilization
                            [   
                                RA-CM3~\cite{racm3}{, }
                                DKA~\cite{dka}{, }
                                UniRAG~\cite{unirag}{, }
                                RMR~\cite{rmr}{, }
                                VisDoM~\cite{visdom}{, }
                                SAM-RAG~\cite{samrag}{, }
                                RAGAR~\cite{ragar}{, }
                                IRAMIG~\cite{iramig}{, }
                                VideoRAG~\cite{videorag}\\
                                RAG-Driver~\cite{ragdriver}{, }
                                VideoRAG~\cite{videorag}{, }
                                Wiki-LLaVA~\cite{wikillava}{, }
                                ReflectiVA~\cite{reflectiva}{, }
                                MMed-RAG~\cite{mmedrag}{, }
                                RAGPT~\cite{ragpt}{, }
                                mRRAG~\cite{mrrag}{, }
                                RULE~\cite{rule}{, }
                                , leaf, text width=46em
                            ]
                        ]
                    ]
                    [
                       Multimodal \\Reasoning,draw=darkyellow,fill=lightyellow,edge+={darkyellow, line width=0.8pt},
                        [   
                            Image \\Reasoning,,draw=darkyellow,fill=lightyellow,edge+={darkyellow, line width=0.8pt},
                            [
                                Vision-Matters~\cite{visionmatters}{, }
                                ReFocus~\cite{refocus}{, }
                                Skywork-R1V~\cite{skyworkr1v}{, }
                                CMMCoT~\cite{cmmcot}{, }
                                Unified-Reward~\cite{cmmcot}{, }
                                MM-RLHF~\cite{mmrlhf}{, }
                                R1-Onevision~\cite{r1onevision}{, }\\
                                LLaVA-CoT~\cite{llavacot}{, }
                                Visual-CoT~\cite{visualcot}{, }
                                LLaVA-Reasoner~\cite{llavareasoner}{, }
                                IXC~\cite{ixc}{, }
                                Insight-V~\cite{insightv}{, }
                                R1-VL~\cite{r1vl}{, }
                                Curr-REFT~\cite{currreft}{, }
                                AR-MCTS~\cite{armcts}{, }\\
                                AURORA~\cite{aurora}{, }
                                VoT~\cite{vot}{, }
                                DCoT~\cite{dcot}{, }
                                Examplar~\cite{examplar}{, }
                                DetToolChain~\cite{dettoolchain}{, }
                                Cantor~\cite{cantor}{, }
                                GoT~\cite{got}{, }
                                CCoT~\cite{ccot}{, }
                                CoTDet~\cite{cotdet}
                                , leaf, text width=46em, ,draw=darkyellow,fill=lightyellow,edge+={darkyellow, line width=0.8pt},
                            ]
                        ]
                        [
                            Video \\Reasoning,draw=darkyellow,fill=lightyellow,edge+={darkyellow, line width=0.8pt},
                            [   
                                IntentQA~\cite{intentqa}{, }
                                AntGPT~\cite{antgpt}{, }
                                VIP~\cite{vip}{, }
                                VioT~\cite{viot}{, }
                                VideoHal~\cite{videohal}{, }
                                CaRDiff~\cite{cardiff}{, }
                                TI-PREGO~\cite{tiprego}{, }
                                ExMRD~\cite{exmrd}{, }
                                CoS~\cite{cos}{, }\\
                                Video-R1~\cite{videor1}{, }
                                VideoRFT~\cite{videorft}{, }
                                UnifiedReward-Vid~\cite{unifiedrewardvideo}{, }
                                VideoMind~\cite{videomind}
                                , leaf, text width=46em,draw=darkyellow,fill=lightyellow,edge+={darkyellow, line width=0.8pt},
                            ]
                        ]
                        [
                            Audio \\Reasoning,draw=darkyellow,fill=lightyellow,edge+={darkyellow, line width=0.8pt},
                            [   
                                Audio-CoT~\cite{audiocot}{, }
                                CoT-ST~\cite{cotst}{, }
                                LPE~\cite{lpe}{, }
                                SpatialSonic~\cite{spatialsonic}{, }
                                Audio-Reasoner~\cite{audioreasoner}{, }
                                Mellow~\cite{mellow}{, }
                                GRPO-Audio~\cite{grpoaudio}
                                , leaf, text width=46em,draw=darkyellow,fill=lightyellow,edge+={darkyellow, line width=0.8pt},
                            ]
                        ]
                    ]
                    [   
                            Multimodal \\Hallucination,draw=darkred,fill=lightred,edge+={darkred, line width=0.8pt},
                            [
                                Analysis,draw=darkred,fill=lightred,edge+={darkred, line width=0.8pt},
                                [
                                    POPE~\cite{pope}{, }
                                    VCDA~\cite{VCD_Ana}{, } VidHalluc~\cite{vidhalluc}{, } Causality~\cite{causality}{, }
                                    Ground~\cite{ground}{, }
                                    Lens~\cite{lens}{, }
                                    F-CLIP~\cite{fclip}{, }
                                    VIC~\cite{vic}{, }
                                    PACU~\cite{pacu}{, }
                                    MAD~\cite{mad}{, }\\
                                    Cantor~\cite{cantor}{, }
                                    Fact~\cite{fact}{, }
                                    PVP~\cite{pvp}{, }
                                    Octopus~\cite{octopus}{, }
                                    VHExpansion~\cite{vhexpansion}{, }
                                    HaELM~\cite{haelm}
                                    , leaf, text width=46em,draw=darkred,fill=lightred,edge+={darkred, line width=0.8pt},
                                ]
                            ]
                            [
                                Detection,draw=darkred,fill=lightred,edge+={darkred, line width=0.8pt},
                                [
                                    DHCP~\cite{dhcp}{, }
                                    VL-Uncertainty~\cite{vluncertainty}{, }
                                    Dentist~\cite{dentist}{, }
                                    Volcano~\cite{volcano}{, }
                                    VFC~\cite{vfc}{, }
                                    HSA-DPO~\cite{hsadpo}{, }
                                    DFTG~\cite{dftg}{, }
                                    FGAIF~\cite{fgaif}{, }
                                    Seeing~\cite{seeing}{, }\\
                                    F-CLIP~\cite{fclip}{, }
                                    LogicCheckGPT~\cite{logiccheckgPT}{, }
                                    Eazy~\cite{eazy}{, }
                                    EFUF~\cite{efuf}{, }
                                    HA-DPO~\cite{hadpo}{, }
                                    HalluciDoctor~\cite{hallucidoctor}{, }
                                    LRV~\cite{lrv}{, }
                                    Octopus~\cite{octopus}
                                    , leaf, text width=46em,draw=darkred,fill=lightred,edge+={darkred, line width=0.8pt},
                                ]
                            ]
                            [
                                Mitigation,draw=darkred,fill=lightred,edge+={darkred, line width=0.8pt},
                                [
                                    RAH~\cite{rah}{, }
                                    BUHR~\cite{buhr}{, }
                                    CATCH~\cite{catch}{, }
                                    VTI~\cite{vti}{, }
                                    ARA~\cite{ara}{, }
                                    BACON~\cite{bacon}{, }
                                    Pelican~\cite{pelican}{, }
                                    AGLA~\cite{agla}{, }
                                    NoiseBoost~\cite{noiseboost}{, }\\
                                    HALC~\cite{halc}{, }
                                    SumGD~\cite{sumgd}{, }
                                    Woodpecker~\cite{woodpecker}{, }
                                    EAGLE~\cite{eagle}{, }
                                    LogicCheckGPT~\cite{logiccheckgPT}{, }
                                    Seeing~\cite{seeing}{, }
                                    F-CLIP~\cite{fclip}{, }
                                    LLaVA-RLHF~\cite{llavarlhf}{, }\\
                                    HDPO~\cite{hdpo}{, }
                                    VDPO~\cite{vdpo}{, }
                                    HELPD~\cite{helpd}{, }
                                    CLIP-DPO~\cite{clipdpo}{, }
                                    REVERIE~\cite{reverie}{, }
                                    CoMT~\cite{comt}{, }
                                    RLAIF-V~\cite{rlaifv}{, }
                                    POVID~\cite{povid}{, }
                                    HACL~\cite{hacl}
                                    , leaf, text width=46em,draw=darkred,fill=lightred,edge+={darkred, line width=0.8pt},
                                ]
                            ]
                    ]
                    [
                       Multimodal \\Safety,draw=darkgreen,fill=lightgreen,edge+={darkgreen, line width=0.8pt},  
                       [    
                            Attack,draw=darkgreen,fill=lightgreen,edge+={darkgreen, line width=0.8pt},  
                            [
                                    AttackLVLM~\cite{attacklvlm}{, }
                                    PM~\cite{pm}{, }
                                    AttackBard~\cite{attackbard}{, }
                                    FigStep~\cite{figstep}{, }
                                    InstructTA~\cite{instructta}{, }
                                    Verbose-Images~\cite{verboseimages}{, }
                                    SGTA~\cite{sgta}{, }
                                    Jailbreaking~\cite{jailbreaking}{, }\\
                                    VisCo~\cite{visco}{, }
                                    Shadowcast~\cite{shadowcast}{, }
                                    BadVLMDriver~\cite{badvlmdriver}{, }
                                    VRP~\cite{vrp}{, }
                                    BMAP~\cite{bmap}{, }
                                    IDEATOR~\cite{ideator}{, }
                                    HIMRD~\cite{himrd}{, }
                                    Badtoken~\cite{badtoken}
                                    , leaf, text width=46em,draw=darkgreen,fill=lightgreen,edge+={darkgreen, line width=0.8pt},  
                            ]
                        ]
                        [    
                            Defense,draw=darkgreen,fill=lightgreen,edge+={darkgreen, line width=0.8pt},  
                            [
                                    DRESS~\cite{dress}{, }
                                    MLLM-Protector~\cite{mllmproctor}{, }
                                    AdaShield~\cite{adashield}{, }
                                    SafeVLM~\cite{safevlm}{, }
                                    MirrorCheck~\cite{mirrorcheck}{, }
                                    CIDER~\cite{cider1}{, }
                                    PIP~\cite{pip1}{, }
                                    VLMGuard~\cite{vlmguard}{, }\\
                                    ETA~\cite{eta}{, }
                                    BlueSuffix~\cite{bluesuffix}{, }
                                    Immune~\cite{immune}{, }
                                    ADPO~\cite{adpo}{, }
                                    SEA~\cite{sea}{, }
                                    VLMGuard-R1~\cite{vlmguardr1}{, }
                                    HoliSafe~\cite{holisafe}
                                    , leaf, text width=46em,draw=darkgreen,fill=lightgreen,edge+={darkgreen, line width=0.8pt},  
                            ]
                        ]
                    ]
                    [
                        Multimodal \\Agents,draw=darkorange,fill=lightorange,edge+={darkorange, line width=0.8pt},  
                        [
                            Closed-source \\Agents,draw=darkorange,fill=lightorange,edge+={darkorange, line width=0.8pt},
                            [
                                CRAFT~\cite{craft}{, }
                                ViperGPT~\cite{vipergpt}{, }
                                CLOVA~\cite{clova}{, }
                                HuggingGPT~\cite{hugginggpt}{, }
                                Chameleon~\cite{chameleon}{, }
                                VisualChatGPT~\cite{visualchatgpt}{, }
                                AssistGPT~\cite{assistgpt}{, }
                                GRID~\cite{grid}{, }\\
                                Drive~\cite{drive}{, }
                                ASSISTGUI~\cite{assistgui}{, }
                                MusicAgent~\cite{musicagent}{, }
                                AudioGPT~\cite{audiogpt}{, }
                                DroidBot-GPT~\cite{droidbot}{, }
                                DEPS~\cite{deps}{, }
                                Cradle~\cite{cradle}{, }
                                Mobile-Agent~\cite{mobileagent}{, }\\
                                SeeAct~\cite{seeact}{, }
                                DoraemonGPT~\cite{doraemongpt}{, }
                                ChatVideo~\cite{chatvideo}{, }
                                AppAgent~\cite{appagent}{, }
                                MM-Navigator~\cite{mmnavigator}{, }
                                WavJourney~\cite{wavjourney}
                                , leaf, text width=46.2em,draw=darkorange,fill=lightorange,edge+={darkorange, line width=0.8pt},
                            ]
                        ]
                        [
                            Open-source \\Agents,draw=darkorange,fill=lightorange,edge+={darkorange, line width=0.8pt},
                            [
                                LLaVA-Interactive~\cite{llavainteractive}{, }
                                JARVIS-1~\cite{jarvis1}{, }
                                STEVE~\cite{steve}{, }
                                EMMA~\cite{emma}{, }
                                MLLM-Tool~\cite{mllmtool}{, }
                                LLaVA-Plus~\cite{llavaplus}{, }
                                GPT4Tools~\cite{gpt4tools}{, }\\
                                WebWISE~\cite{webwise}{, }
                                Auto-GUI~\cite{autogui}
                                , leaf, text width=46em,draw=darkorange,fill=lightorange,edge+={darkorange, line width=0.8pt},
                            ]
                        ]
                    ]
                    [
                       Video \\Perception,draw=darkpurple,fill=lightpurple,edge+={darkpurple, line width=0.8pt},
                       [    
                            Temporal \\Grounding,draw=darkpurple,fill=lightpurple,edge+={darkpurple, line width=0.8pt},
                            [
                                    TFVTG~\cite{zheng2024training}{, }
                                    ChatVTG~\cite{qu2024chatvtg}{, }
                                    Momentor~\cite{qian2024momentor}{, }
                                    VTG-LLM~\cite{guo2025vtg}{, }
                                    E.T.Chat~\cite{liu2024bench}{, }
                                    TRACE~\cite{guo2024trace}{, }
                                    VideoMind~\cite{videomind}{, }
                                    Time-R1~\cite{wang2025time}{, }\\
                                    VideoChat-R1~\cite{li2025videochat}{, }
                                    MUSEG~\cite{luo2025museg}
                                    , leaf, text width=46em,draw=darkpurple,fill=lightpurple,edge+={darkpurple, line width=0.8pt},
                            ]
                        ]
                        [    
                            Dense \\Captioning,draw=darkpurple,fill=lightpurple,edge+={darkpurple, line width=0.8pt},
                            [
                                    VTG-LLM~\cite{guo2025vtg}{, }
                                    E.T.Chat~\cite{liu2024bench}{, }
                                    TRACE~\cite{guo2024trace}{, }
                                    VideoMind~\cite{videomind}{, }
                                    Time-R1~\cite{wang2025time}{, }
                                    VideoChat-R1~\cite{li2025videochat}{, }
                                    MUSEG~\cite{luo2025museg}{, }
                                    VersaVid~\cite{chen2025versavid}{, }\\
                                    VideoCap-R1~\cite{meng2025videocap}{, }
                                    PAM~\cite{lin2025perceive}
                                    , leaf, text width=46em,draw=darkpurple,fill=lightpurple,edge+={darkpurple, line width=0.8pt},
                            ]
                        ] 
                        [    
                            Question \\Answering,draw=darkpurple,fill=lightpurple,edge+={darkpurple, line width=0.8pt},
                            [
                                    MoReVQA~\cite{min2024morevqa}{, }
                                    VersaVid~\cite{chen2025versavid}{, }
                                    DeepVideo-R1~\cite{park2025deepvideo}{, }
                                    VideoGPT+~\cite{maaz2024videogpt+}{, }
                                    TraveLER~\cite{shang2024traveler}{, }
                                    VideoAgent~\cite{videoagent}{, }
                                    MUPA~\cite{dang2025mupa}{, }
                                    PAM~\cite{lin2025perceive}{, }\\
                                    LangRepo~\cite{kahatapitiya2024language}{, }
                                    HierarQ~\cite{azad2025hierarq}{, }
                                    MVU~\cite{ranasinghe2024understanding}{, }
                                    LLoVi~\cite{zhang2023simple}{, }
                                    LVNet~\cite{park2024too}{, }
                                    OmAgent~\cite{zhang2024omagent}{, }
                                    DrVideo~\cite{drvideo}
                                    , leaf, text width=46em,draw=darkpurple,fill=lightpurple,edge+={darkpurple, line width=0.8pt},
                            ]
                        ] 
                    ]
            ]
        \end{forest}
    }
    \caption{A summary of external tools employed to enhance MLLM performance across six challenging tasks.}
    \vspace{-2mm}
    \label{taxo_of_model}
\end{figure*}

\subsubsection{Knowledge Utilization}
After reranking, the remaining knowledge is incorporated into MLLMs to guide response generation. Most methods integrate knowledge in two representative approaches: \textit{training-free approach} and \textit{instruction tuning approach}.

\noindent \textbf{Training-free approach.}
Training-free approach, particularly In-Context Learning (ICL), incorporates the retrieved knowledge without requiring parameter updates. It treats the retrieved content as contextual examples to steer MLLM generation. 
For instance, RA-CM3~\cite{racm3} extends RAG into both image and text generation with in-context learning.
DKA~\cite{dka} utilizes the retrieved samples and knowledge as few-shot examples to enhance model performance on the knowledge-based visual question answering task.
Similarly, UniRAG~\cite{unirag} employs the retrieved data as in-context examples for the caption generation task.
RMR~\cite{rmr} employs a bi-modal retrieval module to identify the most relevant question-answer pairs and encourages the model to engage with the reasoning processes inherent in the retrieved content. 
While VisDoM~\cite{visdom} and SAM-RAG~\cite{samrag} employ Chain of Thought (CoT) reasoning with the retrieved knowledge to enhance MLLM accuracy on multimodal reasoning.
RAGAR~\cite{ragar} introduces the chain of RAG and the tree of RAG to further improve the MLLM's ability to utilize the retrieved knowledge for fact checking.
IRAMIG~\cite{iramig} comprises modality-specific retrieval modules and a fusion component that synthesizes knowledge, ultimately enriching the model’s understanding and output generation.
RAG-Driver~\cite{ragdriver} introduces RAG into the autonomous driving domain to enhance the explainability of MLLMs as a driving agent.
VideoRAG~\cite{videorag} incorporates the extracted knowledge into an existing MLLM as auxiliary information, alongside video frames and queries, in a plug-and-play manner to generate responses for the video question answering.

\noindent \textbf{Instruction Tuning-based approach.}
Since many MLLMs are not pretrained to effectively consume in-context knowledge, the instruction tuning approach explicitly fine-tunes MLLMs to better leverage retrieved content.
For example, Wiki-LLaVA~\cite{wikillava} empowers LLaVA~\cite{liu2023visual} with retrieved passages and fine-tunes the model to augment its capabilities of exploiting retrieved passages.
ReflectiVA~\cite{reflectiva} introduces a reflective token to determine the need for knowledge retrieval to enable the MLLM to manage external knowledge while preserving fluency.
MMed-RAG~\cite{mmedrag} proposes a provable RAG-based preference fine-tuning strategy to adapt MLLMs and RAG to the medical domain.
RAGPT~\cite{ragpt} fine-tunes a missing modality generator to handle the retrieved knowledge with incomplete modalities, and trains a prompter to capture contextual knowledge from relevant instances.
mRRAG~\cite{mrrag} introduces the Retrieval-Reflection process to distinguish different user queries and avoid redundant retrieval calls.
RULE~\cite{rule} curates a preference dataset to fine-tune the model to balance its dependence on inherent and retrieved knowledge for generation.

\subsection{Multimodal Reasoning}
By extending the benefits of Chain-of-Thought reasoning to multimodal contexts, multimodal reasoning has garnered increasing research interest, especially in the realm of MLLMs. In this section, we review how external tools can enhance the reasoning capabilities of MLLMs across different modalities, including \textit{image}, \textit{video}, and \textit{audio}.

\subsubsection{Image Reasoning}
The rapid growth of image data has accelerated the development of MLLMs for complex reasoning tasks such as mathematical and spatial reasoning. Recent studies have explored a variety of model designs, data strategies, and training paradigms to improve MLLMs' reasoning capabilities.

To integrate rich visual semantics into reasoning, several studies explore various novel ways to inject visual cues and trace visual thoughts.
For instance, Vision-Matters~\cite{visionmatters} proposes three perturbation-based distractors to incorporate more discriminative visual signals, improving MLLM performance on mathematical reasoning tasks.
Similarly, ReFocus~\cite{refocus} introduces Python-based visual editing tools, enabling MLLMs to perform visual thought—iterative image modifications that support visual reasoning.
Extending this idea, MVoT~\cite{mvot} allows models to visualize internal reasoning steps, providing interpretable visual reasoning traces and excelling in spatial tasks.

To transfer strong language reasoning abilities to the visual domain, Skywork-R1V~\cite{skyworkr1v} introduces an MLP adapter for handling variable-length multimodal inputs.
CMMCoT~\cite{cmmcot} proposes test-time memory augmentation to boost reasoning without increasing parameters, while Unified-Reward~\cite{cmmcot} further enhances generalization by curating a unified reward dataset for image generation and understanding.
Insight-V~\cite{insightv} explores modular architectures, combining agents or specialized components (\eg, Interactor, Explorer) for better spatial and temporal reasoning.

Many recent efforts focus on constructing high-quality reasoning data for MLLM training.
Large-scale supervision signals are generated using state-of-the-art models such as GPT-4o~\cite{gpt4o} in MM-RLHF~\cite{mmrlhf}, R1-Onevision~\cite{r1onevision}, Visual-CoT~\cite{visualcot}, LLaVA-CoT~\cite{llavacot}. These datasets offer large-scale, diverse, and high-quality reasoning examples with fine-grained annotations.
Building on these datasets, LLaVA-Reasoner~\cite{llavareasoner} uses GPT-4o~\cite{gpt4o} to generate rationales from minimal supervision, showing strong generalization to direct-answer reasoning tasks.

Reward modeling plays a key role in guiding MLLM reasoning.
For example, IXC~\cite{ixc} introduces a lightweight multimodal reward model with a score head that supervises MLLM training.
R1-VL~\cite{r1vl} leverages GPT-4o~\cite{gpt4o} to extract key reasoning steps and applies step-wise rewards to reduce sparsity and encourage logical consistency.
Similarly, Curr-REFT~\cite{currreft} adopts curriculum-based reinforcement finetuning using GPT-4o~\cite{gpt4o} as the reward model, enabling staged learning from visual recognition to complex reasoning.

To embed explicit knowledge and structure into reasoning, a range of graph-based CoT approaches have been explored. For example, GoT~\cite{got} constructs a Graph-of-Thought, treating thought units as nodes with logical dependencies as edges, which enhances scientific reasoning.
CCoT~\cite{ccot} and CoTDet~\cite{cotdet} incorporate structured representations like scene graphs, hypergraphs, and knowledge graphs into the CoT process, guiding MLLMs to extract relevant knowledge and improve benchmark performance.
Cantor~\cite{cantor} further mitigates hallucinations by utilizing external engines in multimodal CoT reasoning.

Retrieval and in-context demonstration improve inference by grounding models in relevant examples.
AR-MCTS~\cite{armcts} moves beyond beam search by dynamically retrieving key reasoning insights per step to expand the reasoning space.
DCoT~\cite{dcot} and Examplar~\cite{examplar} retrieve optimal in-context examples from curated clusters, enhancing inference accuracy and generalization.
DetToolChain~\cite{dettoolchain} proposes a CoT prompting toolkit to help MLLMs focus on regional visual features, improving detection performance.
These advancements in MLLMs demonstrate a multifaceted push toward more interpretable, generalizable, and domain-adaptive reasoning systems. Through innovations in visual representation, modular architecture, reward modeling, data curation, and structured reasoning, the field evolves toward more robust and human-aligned intelligence.

\subsubsection{Video Reasoning}
Compared to image reasoning, video reasoning poses significantly greater challenges due to its inherent temporal complexity, particularly in long-duration or event-dense scenarios. To address these issues, recent research has focused on improving temporal modeling, introducing structured representations, and designing advanced prompting strategies to support MLLMs in video understanding.

One core challenge in video reasoning lies with modeling temporal dynamics across frames. IntentQA~\cite{intentqa} tackles this by employing a dynamic graph Transformer to extract region-based temporal graphs, which are then passed through a commonsense reasoning module to enhance contextual understanding over time. AntGPT~\cite{antgpt} abstracts video content into action sequences, enabling an LLM to infer goals and temporal relations and achieve strong generalization across benchmarks. TI-PREGO~\cite{tiprego} introduces a two-stage architecture that combines action recognition with CoT-based reasoning to predict future actions, offering enhanced interpretability in sequential event modeling. Meanwhile, models such as Video-R1~\cite{videor1}, UnifiedReward-Vid~\cite{unifiedrewardvideo}, and VideoRFT~\cite{videorft} leverage powerful MLLMs to automatically generate CoT annotations, facilitating scalable and effective training for video reasoning tasks.
To improve video understanding, several studies incorporate structured representations. For example, VioT~\cite{viot} introduces a scene graph encoder that guides the model from low-level visual perception to high-level semantics, enhancing spatio-temporal reasoning. CaRDiff~\cite{cardiff} combines CoT prompting with saliency maps generated by GroundingDINO~\cite{groundingdino} and Stable Diffusion~\cite{stablediffusion}, grounding video frames with spatially explicit information. ExMRD~\cite{exmrd} integrates GPT-4~\cite{gpt4o} to refine video descriptions, retrieve external knowledge, and perform CoT-based inference, achieving state-of-the-art results in micro-video rumor detection.

Another typical approach emphasizes content-aware reasoning through selective frame processing and targeted CoT strategies. For example, VideoHal~\cite{videohal} selects key frames based on question-frame similarity using CLIP~\cite{clip}, and combines CoT prompting with in-context learning to reduce hallucination in video QA. CoS~\cite{cos} introduces Chain-of-Shot prompting, where an MLLM selects representative shots to construct a coherent temporal context for long-video understanding. VIP~\cite{vip} further advances this by using GPT-4V\cite{gpt4o} to extract scene- and object-level summaries and proposes a new inference-time dataset to evaluate CoT-based video reasoning. 
VideoMind~\cite{videomind} proposes a Chain-of-LoRA framework that supports dynamic role-switching via lightweight LoRA adapters, balancing reasoning efficiency with model adaptability without incurring the cost of multiple models.
These advances demonstrate the crucial role of chain-of-thought reasoning in video understanding. By addressing challenges in temporal modeling, semantic grounding, and efficient inference, these methods collectively push the boundaries of multimodal reasoning in complex video-based tasks.

\subsubsection{Audio Reasoning}
Multimodal reasoning has been effectively extended to step-wise and structured audio processing, bridging the gap between raw waveform signals and language semantics.

Audio-CoT~\cite{audiocot} pioneers the integration of CoT reasoning into large audio-language models by leveraging few-shot examples to guide the reasoning process, significantly improving performance on tasks of low to moderate difficulty.
CoT-ST~\cite{cotst} introduces a multimodal speech translation model that decomposes the task into two sequential steps—speech recognition and translation—using CoT reasoning to enhance interpretability and performance.
LPE~\cite{lpe} adopts CoT prompting alongside a multiscale adapter to generate empathetic responses by grounding on both spoken content and inferred emotional cues.
SpatialSonic~\cite{spatialsonic} applies CoT prompting to infer sounding objects and their attributes, while incorporating a spatial-aware encoder and azimuth state matrices to guide the generation of immersive and controllable spatial audio from text.

To further enhance reasoning capacity, Audio-Reasoner~\cite{audioreasoner} employs Gemini~\cite{gemini} for secondary labeling via structured CoT, introducing a large-scale model capable of deep auditory reasoning.
Mellow~\cite{mellow} improves alignment between audio and language modalities by using Amazon Mechanical Turk for annotation and introducing dual audio mappers, achieving state-of-the-art results among small-scale audio-language models.
GRPO-Audio~\cite{grpoaudio} explores the use of Group Relative Policy Optimization for training large-scale audio-language models, demonstrating the potential of reinforcement learning in this domain.
These advancements collectively highlight the versatility and effectiveness of multimodal chain-of-thought reasoning in audio domains.

\subsection{Multimodal Hallucination}
Multimodal hallucination~\cite{hallucinationsurvey,xing2024mitigating,zhang2024seeing} refers to the phenomenon where the textual responses generated by MLLMs fail to align with the corresponding visual input. This misalignment presents a significant barrier to the practical deployment of MLLMs and raises concerns about their reliability in real-world applications. As a fundamental problem, multimodal hallucination has garnered increasing research attention. In this section, we review how external tools can assist MLLMs in addressing hallucination challenges from three key perspectives: \textit{Hallucination Cause Analysis}, \textit{Hallucination Detection}, and \textit{Hallucination Mitigation}.

\subsubsection{Hallucination Cause Analysis}
Understanding the underlying causes of hallucinations is essential for identifying and mitigating hallucinations. In this section, we explore how external tools can facilitate a deeper analysis of hallucination sources. Most existing work can be broadly categorized into two representative approaches: \textit{grounding approach} and \textit{cognitive approach}.

\noindent \textbf{Grounding Approach.}
The grounding approach analyzes the hallucination causes by investigating the alignment between the generated content and the grounded objects in the image. For example, POPE~\cite{pope} explores the causes of multimodal hallucinations by analyzing object co-occurrence and object frequency with the help of the segmentation model SEEM~\cite{seem}.
VCDA~\cite{VCD_Ana} investigates the impact of various image augmentation strategies on mitigating multimodal hallucination. Specifically, by introducing diffusion noise~\cite{ho2020denoising}, applying down-sampling, and leveraging InstructPix2Pix~\cite{brooks2023instructpix2pix} for image editing, the authors demonstrate that fusing predictions across these augmented versions can significantly reduce hallucinations.
VidHalluc~\cite{vidhalluc} extracts saliency maps using DINO v2~\cite{oquab2023dinov2} to identify the source of video hallucinations and reweights the corresponding visual features to mitigate their impact.
Ground~\cite{ground} systematically evaluates the influence of fine-grained object grounding on hallucination in LVLMs, with the FaithScore used to quantify performance.
Lens~\cite{lens} examines multimodal hallucinations through attention maps over visual tokens and derives actionable findings to detect and reduce hallucinations.
F-CLIP~\cite{fclip} investigates the distributions of hallucinated and non-hallucinated outputs by extracting nouns using the spaCy parser and comparing their visual-textual alignment.

\noindent\textbf{Cognitive Approach.}
The cognitive approach analyzes the hallucination causes by investigating the reasoning and decision-making processes of MLLMs. From a causal perspective, Causality~\cite{causality} investigates multimodal hallucinations by performing image interventions with Grounding DINO~\cite{groundingdino} and IA~\cite{yu2023inpaint}, and further applies embedding interventions using edited embeddings retrieved from an auxiliary database.
VIC~\cite{vic} explores hallucinations through the lens of CoT reasoning by generating visual inference chains with LLMs such as GPT-4~\cite{gpt4o}.
PACU~\cite{pacu} leverages existing LLMs~\cite{gpt4o} to automatically augment prompts and evaluate their impact on multimodal hallucination.
MAD~\cite{mad} frames hallucination as a complex reasoning task and adopts a multi-agent debate setup with Gemini~\cite{gemini} to analyze the underlying reasoning patterns of MLLMs.
Cantor~\cite{cantor} investigates multimodal CoT and addresses hallucination by utilizing GPT~\cite{gpt4o} and Gemini~\cite{gemini} as external reasoning engines.
Fact~\cite{fact} examines hallucinations from a code-centric perspective, using a program generator~\cite{gpt4o} to produce multimodal rationales that are faithful, concise, and transferable.
PVP~\cite{pvp} probes hallucinations through counterfactual reasoning, employing self-generated counterfactual keywords with GPT-4~\cite{gpt4o}.
VHExpansion~\cite{vhexpansion} explores hallucination causes by perturbing both the question and the answer with negations generated by ChatGPT~\cite{gpt4o}.
Finally, HaELM~\cite{haelm} applies LLaMA to detect and interpret hallucinations in MLLMs, with the additional benefits of low cost, reproducibility, and privacy preservation.

\subsubsection{Hallucination Detection}

Accurately detecting hallucinated content is crucial for effective mitigation, as it enables targeted correction while avoiding modification of correct responses. In this section, we review how prior works leverage external tools to identify hallucinations from two perspectives: \textit{Detector-based approach} and \textit{LLM-based approach}.

\noindent \textbf{Detector-based approach.}
This kind of approach introduces external tools or trainable modules to detect hallucinations. For instance, DHCP~\cite{dhcp} introduces two learnable detectors to identify multimodal hallucinations by analyzing cross-modal attention patterns between hallucinated and non-hallucinated states.
Seeing~\cite{seeing} and F-CLIP~\cite{fclip} propose training-free solutions that use CLIP~\cite{clip} to guide the decoding process and reduce hallucinations.
Eazy~\cite{eazy} extracts object tokens through POS tagging and dependency parsing to identify hallucinated elements.
EFUF~\cite{efuf} computes CLIP~\cite{clip}-based image-relevance scores to construct datasets for hallucination detection.
HalluciDoctor~\cite{hallucidoctor} employs the BERT-based BEM metric to evaluate response consistency and detect hallucinations.
Finally, Octopus~\cite{octopus} introduces a dedicated decision token and classifier to identify hallucination types at each decoding step.

\noindent \textbf{LLM-based approach.}
This kind of approach introduces LLMs or MLLMs to detect hallucinations. For instance, VL-Uncertainty~\cite{vluncertainty} applies semantic-equivalent perturbations to both visual and textual prompts, then leverages LLMs~\cite{gpt4o} to rephrase the original queries and detect inconsistencies.
Dentist~\cite{dentist} and Volcano~\cite{volcano} utilize ChatGPT~\cite{chatgpt} to classify hallucinations and revise responses in a prompt-driven manner.
VFC~\cite{vfc} reconstructs images using Stable Diffusion XL~\cite{stablediffusion} and evaluates image–caption consistency with GPT-4~\cite{gpt4o}.
HSA-DPO~\cite{hsadpo} uses GPT-4V~\cite{gpt4o} to produce fine-grained, sentence-level AI feedback and integrates this into Direct Preference Optimization (DPO) for hallucination mitigation.
DFTG~\cite{dftg} extracts essential information from model responses and images with ChatGPT~\cite{chatgpt} and Grounding DINO~\cite{groundingdino} to detect hallucinations.
FGAIF~\cite{fgaif} predicts hallucination types for each segment using GPT~\cite{gpt4o} and incorporates fine-grained feedback into Proximal Policy Optimization (PPO) to iteratively improve multimodal responses.
LogicCheckGPT~\cite{logiccheckgPT} checks logical consistency by prompting ChatGPT~\cite{chatgpt} with targeted questions derived from the logical structure of the response.
HA-DPO~\cite{hadpo} frames hallucination detection as a preference-selection task and utilizes ChatGPT~\cite{chatgpt} to identify and mitigate hallucinated content.
LRV~\cite{lrv} ranks alternative responses with GPT-4~\cite{gpt4o} to filter out hallucinated candidates.

\subsubsection{Hallucination Mitigation}

This section presents a comprehensive review of tool-enhanced approaches for mitigating hallucinations in MLLMs. Specifically, three representative approaches have been explored with different design principles and usage of external tools, namely, \textit{Input-augmented approach}, \textit{Training-based approach}, and \textit{Calibration-based approach}.

\noindent \textbf{Input-augmented Approach.}
Input-augmented approach enhances inference by supplying MLLMs with additional information (\eg, visual cues or retrieved content) from external tools. For example, RAH~\cite{rah} utilizes SAM~\cite{sam} to provide segmentation masks that help MLLMs identify context-relevant objects.
BUHR~\cite{buhr} augments MLLMs with object and attribute verification via Grounding DINO~\cite{groundingdino}.
CATCH~\cite{catch} applies SAM~\cite{sam} to segment foreground and background elements and integrates these visual cues at the logit level to improve response grounding.
ARA~\cite{ara} retrieves external captions with CLIP~\cite{clip} to enrich the MLLM’s context and reduce hallucinated content.
BACON~\cite{bacon} generates candidate object regions with Grounding DINO~\cite{groundingdino}, then selects the most relevant descriptions using CLIP~\cite{clip}.
AGLA~\cite{agla} matches the input image and query to isolate query-relevant image regions, masking out distractions and yielding more accurate responses.
NoiseBoost~\cite{noiseboost} introduces noise perturbations on pre-trained visual features to encourage a more balanced attention distribution.

\noindent \textbf{Training-based Approach.}
Training-based approach leverages high-quality data, curated or generated by external tools, to fine-tune MLLMs and strengthen their robustness against hallucinations. For example, EAGLE~\cite{eagle} employs a segmentation model~\cite{sam} to provide segmentation masks for MLLMs and improve the grounding ability of the image encoder through GaLore training~\cite{galore}. 
LLaVA-RLHF~\cite{llavarlhf} utilizes GPT-4~\cite{gpt4o} to generate training data and adapts Reinforcement Learning from Human Feedback (RLHF) to the multimodal domain by augmenting it with additional factual information, which can improve the general capabilities and mitigate hallucinations for MLLMs.
HDPO~\cite{hdpo}, POVID~\cite{povid}, and HACL~\cite{hacl} utilize GPT-4~\cite{gpt4o} to generate negative samples and perform Direct Preference Optimization (DPO)~\cite{dpo} to mitigate hallucinations for MLLMs.
Similarly, VDPO~\cite{vdpo} employs ChatGPT~\cite{chatgpt} to replace ground-truth objects in the responses and utilizes diffusion models~\cite{ddpm} to generate noisy images. Based on the generated negative samples, the authors perform vision-guided DPO~\cite{dpo} to mitigate hallucinations for MLLMs.
HELPD~\cite{helpd} prompts GPT-4~\cite{gpt4o} to synthesize a longer caption for each image based on the short captions and then perform fine-tuning to improve the text generation quality of MLLMs.
CLIP-DPO~\cite{clipdpo} utilizes CLIP~\cite{clip} to rank pre-generated responses and select high-quality responses for DPO~\cite{dpo} training.
REVERIE~\cite{reverie} employs GeminiVision-Pro~\cite{gemini} to annotate the instructions, responses, and rationales for each image. Based on the generated high-quality data, the authors perform instruction tuning to enhance the model's reasoning proficiency.
CoMT~\cite{comt} employs ChatGPT~\cite{chatgpt} to imitate the cognitive process of human doctors and generate the chain of thought for model training, enhancing the inferential ability during diagnosis and alleviating hallucination problems.
RLAIF-V~\cite{rlaifv} introduces an additional MLLM as a reference model and performs self-feedback guidance for inference-time scaling, substantially enhancing the trustworthiness of MLLMs.

\noindent \textbf{Calibration-based Approach.}
Calibration-based approach post-processes and revises hallucinated outputs to improve factual consistency. For example, VTI~\cite{vti} employs ChatGPT~\cite{chatgpt} to generate hallucinated responses and then compute the textual and visual direction with less hallucination. By adding the direction to the latent states for new samples, MLLMs can generate less-hallucinated responses.
Pelican~\cite{pelican} utilizes ChatGPT~\cite{chatgpt} to decompose and validate the generated responses, and then integrates the verification into final responses to mitigate hallucinations in the original responses.
HALC~\cite{halc} applies Grounding DINO~\cite{groundingdino} and OWL v2~\cite{owlv2} to locate the important visual tokens within the original image input and utilizes the located tokens to calibrate the original outputs with contrastive decoding~\cite{cd}.
Woodpecker~\cite{woodpecker} utilizes ChatGPT~\cite{chatgpt} to extract key concepts from the responses of MLLM and then validate the extracted textual information with the visual information extracted with Grounding DINO~\cite{groundingdino}. Guided by the validation results, ChatGPT~\cite{chatgpt} can act as a corrector and modify the hallucinations in the original responses.
Seeing~\cite{seeing} and F-CLIP~\cite{fclip} calibrate the responses of MLLMs by using the image-text similarities computed by CLIP~\cite{clip}.
SumGD~\cite{sumgd} employs Part-of-Speech (POS) tagging to detect objects in responses and revise hallucinated objects with Flan-T5~\cite{t5}.
LogicCheckGPT~\cite{logiccheckgPT} utilizes ChatGPT~\cite{chatgpt} to revise the hallucinated contents in the generated responses, thus increasing the reliability of MLLMs.

\subsection{Multimodal Safety}

While MLLMs offer substantial benefits and have markedly improved user experiences across a wide range of applications, ensuring their safety and security remains a critical concern. In this section, we review how external tools have been leveraged to enhance the safety of MLLMs from two perspectives: \textit{multimodal attacks} and \textit{multimodal defenses}.

\subsubsection{Multimodal Attack}
Extensive research has explored attack strategies targeting MLLMs, as identifying MLLM vulnerabilities is crucial for developing robust defenses and enhancing model safety. This section reviews how prior works use external tools to craft adversarial samples from two aspects: \textit{image-based attacks} and \textit{prompt-based attacks}.

\noindent \textbf{Image-based attacks.} Image-based attacks typically employ image editing techniques to manipulate input images and generate adversarial or backdoored samples. For instance,
AttackLVLM~\cite{attacklvlm} and IDEATOR~\cite{ideator} use the text-to-image model Stable Diffusion~\cite{stablediffusion} to create targeted adversarial images.
AttackBard~\cite{attackbard} utilizes CLIP~\cite{clip} as surrogate models to conduct image embedding attacks.
Badtoken~\cite{badtoken} introduces a teacher model to guide backdoored image embeddings, enabling flexible and stealthy token-level modifications to the model outputs.
Verbose-Images~\cite{verboseimages} crafts imperceptible perturbations that increase MLLMs' energy latency cost, leading to excessive energy and computation usage.
VisCo~\cite{visco} defines a visual-centric jailbreak setting using InternVL2.5~\cite{internvl25} and Stable Diffusion~\cite{stablediffusion} to generate prompts that reliably trigger harmful outputs from black-box MLLMs.
BadVLMDriver~\cite{badvlmdriver} applies InstructPix2Pix~\cite{brooks2023instructpix2pix} for image editing and performs visual instruction tuning to optimize adversarial effectiveness.
VRP~\cite{vrp} employs LLMs to generate high-risk character descriptions and uses Stable Diffusion~\cite{stablediffusion} to synthesize corresponding adversarial images.

\noindent \textbf{Prompt-based attacks.} Prompt-based attacks modify the input prompts to deceive MLLMs. For example,
PM~\cite{pm} proposes a general-purpose dataset for prompt injection and uses behavior matching to construct four types of hijacking attacks, achieving high success rates.
FigStep~\cite{figstep}, SGTA~\cite{sgta}, and InstructTA~\cite{instructta} use GPT-4~\cite{gpt4o} to generate rephrased jailbreaking prompts that embed harmful instructions.
Jailbreaking~\cite{jailbreaking} introduces de-embedding and de-tokenization techniques to derive textual jailbreak prompts from image-based ones, demonstrating high efficiency and cross-model transferability.
Shadowcast~\cite{shadowcast} leverages a pre-trained VLM~\cite{liu2023visual} to generate captions, which are then paraphrased by GPT-3.5~\cite{chatgpt} to emphasize harmful content and craft poisoned images.
BMAP~\cite{bmap} uses ChatGPT~\cite{chatgpt} to diagnose failed jailbreak attempts and iteratively refine prompts for improved attack effectiveness.
HIMRD~\cite{himrd} utilizes an auxiliary LLM to decompose risk into textual and visual components, guiding the victim model to produce harmful outputs through iterative prompting until success or a predefined limit is reached.

\subsubsection{Multimodal Defense}
Defense strategies are critical for enhancing the robustness of MLLMs against a wide range of adversarial threats, particularly in high-stakes domains such as healthcare and autonomous driving. This section reviews how prior works use external tools to defend adversarial samples from two aspects: \textit{attack detection} and \textit{attack mitigation}.

\noindent \textbf{Attack detection.} Prior work on attack detection typically employs external tools to identify adversarial inputs. For instance, 
MLLM-Protector~\cite{mllmproctor} introduces a lightweight harm detector that first assesses whether a response is harmful, followed by a detoxifier module that corrects harmful content. DRESS~\cite{dress} leverages GPT-4~\cite{gpt4o} to provide evaluative feedback on MLLM-generated responses, identifying their strengths and weaknesses and aligning outputs with human preferences to enhance robustness. 
MirrorCheck~\cite{mirrorcheck} utilizes Text-to-Image (T2I) models to generate visual representations based on captions from target MLLMs, then measures the embedding similarity between the input and generated images to detect adversarial samples. 
CIDER~\cite{cider1} incorporates a diffusion-based denoiser to preprocess visual inputs, enabling differentiation between clean and adversarial images for effective jailbreak detection. 
PIP~\cite{pip1} introduces lightweight classifiers (e.g., SVM, MLP) that detect adversarial examples by analyzing attention maps produced by MLLMs. 
ETA~\cite{eta} and VLMGuard~\cite{vlmguard} employ CLIP~\cite{clip} in conjunction with MLPs to assess the safety of visual content by identifying unsafe semantic elements. SEA~\cite{sea} uses GPT-4~\cite{gpt4o} to identify and replace harmful phrases, generating detoxified outputs to mitigate attacks. 
HoliSafe~\cite{holisafe} proposes a safety-focused MLLM architecture incorporating a learnable image safety meta token and a dedicated safety head.

\noindent \textbf{Attack mitigation.} For attack mitigation, many approaches incorporate external tools to train MLLMs or rewrite their responses. 
For example, AdaShield~\cite{adashield} introduces a defense prompt generator based on LLMs, which crafts prompts to protect MLLMs from malicious inputs. 
SafeVLM~\cite{safevlm} utilizes GPT-4~\cite{gpt4o} to create a visual safety alignment dataset and integrates three safety modules to enhance overall safety performance. BlueSuffix~\cite{bluesuffix} proposes a two-part solution: a diffusion-based image purifier for sanitizing jailbreak images and an LLM-based text purifier for rewriting adversarial prompts. 
Immune~\cite{immune} employs a safety reward model with controlled decoding mechanisms to defend against jailbreak attacks during inference. 
ADPO~\cite{adpo} introduces an adversarially trained reference model and an adversarial-aware DPO loss to enforce safety alignment, maintaining MLLM robustness against sophisticated attacks. 
VLMGuard-R1~\cite{vlmguardr1} enhances user inputs through a reasoning-guided rewriter, which dynamically interprets text-image interactions to generate refined prompts that improve safety across a range of MLLMs.

\subsection{Multimodal Agents}
A multimodal agent is a system capable of perceiving information from a multimodal environment, reasoning and planning over complex tasks, and making informed decisions to execute appropriate actions. By enabling continuous interaction with the multimodal environment, such agents extend the capabilities of MLLMs to perform a broad range of real-world tasks. In this section, we review how external tools have been employed to enhance the functionality of two major categories of multimodal agents: \textit{closed-source agents} and \textit{open-source agents}.

\subsubsection{Closed-source Agents}
Closed-source multimodal agents rely on proprietary LLMs or MLLMs, such as ChatGPT~\cite{chatgpt} and GPT-4V~\cite{gpt4o}, as core planners or reasoners. These agents typically employ prompt-based techniques to guide decision-making and task planning. 
One typical scenario leverages external tools to empower MLLMs to handle image-based tasks. For example,
CRAFT~\cite{craft} equips LLMs with a curated toolset for complex multimodal challenges, allowing dynamic retrieval of tools to solve tasks. 
ViperGPT~\cite{vipergpt} composes MLLMs into callable subroutines using code generation, integrating tools like GLIP~\cite{glip}, ChatGPT~\cite{chatgpt}, and X-VLM~\cite{xvlm}. 
CLOVA~\cite{clova} generates and executes tool-specific programs within a closed-loop learning framework, enabling tool updates for new environments. 
HuggingGPT~\cite{hugginggpt} connects diverse community models (e.g., via Hugging Face) to handle AI tasks collaboratively. 
Chameleon~\cite{chameleon} synthesizes programs by combining LLMs, vision models, search engines, and heuristics for complex reasoning.
Other agents, like VisualChatGPT~\cite{visualchatgpt}, integrate Visual Foundation Models (VFMs) to enable perception, generation, and editing of images. AssistGPT~\cite{assistgpt} introduces a PEIL (Plan, Execute, Inspect, Learn) framework for question answering. 

Beyond image tasks, multimodal agents are increasingly applied across other modalities and real-world domains. 
Audio-based agents like MusicAgent~\cite{musicagent}, WavJourney~\cite{wavjourney}, and AudioGPT~\cite{audiogpt} integrate music and speech tools into autonomous pipelines to enrich user experiences.
Video-oriented agents such as DoraemonGPT~\cite{doraemongpt} and ChatVideo~\cite{chatvideo} manage and query video properties through database-driven interactions. 
GUI and mobile agents like ASSISTGUI~\cite{assistgui}, DroidBot-GPT~\cite{droidbot}, AppAgent~\cite{appagent}, and Mobile-Agent~\cite{mobileagent} automate app interactions by parsing GUI states into LLM prompts. 
Web and robotics agents include SeeAct~\cite{seeact} for web-based actions, GRID~\cite{grid} for robotic skill learning and adaptation, and MM-Navigator~\cite{mmnavigator} for smartphone GUI navigation using GPT-4V.
Other notable systems include DEPS~\cite{deps}, which enables robust multi-task completion in Minecraft through error-aware plan correction; Drive~\cite{drive}, which showcases closed-loop autonomous driving; and Cradle~\cite{cradle}, a general computer control agent that converts screen inputs into executable actions via high-level planning and low-level command execution.

\subsubsection{Open-source Agents}
In addition to employing closed-source models as agents, several studies have explored fine-tuning open-source models to endow them with capabilities for decision-making, planning, and tool invocation.
For example, LLaVA-Interactive~\cite{llavainteractive} demonstrates a research prototype for multimodal human-AI interaction by combining three capabilities—visual dialogue via LLaVA~\cite{liu2023visual}, image segmentation from SEEM~\cite{seem}, and image generation/editing from GLIGEN~\cite{gligen}—without additional model training.
JARVIS-1~\cite{jarvis1} presents an open-world agent for the Minecraft environment, capable of perceiving multimodal inputs, generating complex plans, and performing embodied control.
Similarly, STEVE~\cite{steve} integrates three core modules—vision perception, language understanding, and code-based action execution—to enable agents to operate effectively within Minecraft.
EMMA~\cite{emma} introduces a novel DAgger-DPO training algorithm that enables rapid adaptation to dynamic visual environments and generalization across a wide range of tasks without explicit supervision.
MLLM-Tool~\cite{mllmtool} enhances open-source LLMs with multimodal encoders and a tool selection mechanism, allowing the model to interpret multimodal instructions and invoke appropriate tools for complex tasks.
LLaVA-Plus~\cite{llavaplus} extends LLaVA into a general-purpose multimodal assistant capable of handling diverse multimodal inputs, activating tools as needed, and dynamically composing their outputs to complete real-world tasks.
GPT4Tools~\cite{gpt4tools} applies the self-instruct paradigm to enable open-source LLMs to leverage external tools for visual reasoning and image generation.
WebWISE~\cite{webwise} employs in-context learning to automate web software tasks—such as clicking, scrolling, and text input—achieving competitive performance without the need for large-scale demonstrations or trial-and-error training.
Finally, Auto-GUI~\cite{autogui} proposes a chain-of-action framework that utilizes a combination of past actions and future action plans to guide decision-making in GUI-based environments.

\definecolor{hidden-draw}{RGB}{205, 44, 36}
\definecolor{hidden-orange}{RGB}{243,202,120}
\definecolor{hidden-blue}{RGB}{194,232,247}
\definecolor{hidden-yellow}{RGB}{242,244,193}
\definecolor{tree-level-1}{RGB}{245,20,85}
\definecolor{tree-level-2}{RGB}{246,86,118}
\definecolor{tree-level-3}{RGB}{248,177,193}
\definecolor{tree-leaf}{RGB}{176,230,198}
\definecolor{Self}{RGB}{255,0,128}
\definecolor{Ensemble}{RGB}{0,127,255}
\definecolor{Iterative}{RGB}{153,51,255}
\definecolor{exemplar1}{RGB}{136,98,148}
\definecolor{exemplar2}{RGB}{148,210,242}
\definecolor{knowledge1}{RGB}{249,219,152}
\definecolor{knowledge2}{RGB}{255,245,220}

\definecolor{darkblue}{HTML}{000099}
\definecolor{lightblue}{HTML}{D1EEFF}

\definecolor{rootdarkblue}{HTML}{000066}
\definecolor{rootlightblue}{HTML}{C6F3FF}
\definecolor{darkblue}{HTML}{000099}
\definecolor{lightblue}{HTML}{D1EEFF}
\definecolor{darkyellow}{HTML}{FF9B05}
\definecolor{lightyellow}{HTML}{FAFFDD}
\definecolor{darkred}{HTML}{CC0000}
\definecolor{lightred}{HTML}{FFE2DD}
\definecolor{darkgreen}{HTML}{006600}
\definecolor{lightgreen}{HTML}{F3FFE5}
\definecolor{darkorange}{HTML}{FF8000}
\definecolor{lightorange}{HTML}{FFFCD3}
\definecolor{darkpurple}{HTML}{4C0099}
\definecolor{lightpurple}{HTML}{E6D8FF}

\tikzstyle{my-box}=[
    rectangle,
    draw=darkblue,
    rounded corners,
    text opacity=1,
    minimum height=1.5em,
    minimum width=5em,
    inner sep=2pt,
    align=center,
    fill opacity=.2,
]
\tikzstyle{leaf}=[my-box, minimum height=1.5em,
    fill=lightblue, text=black, align=left,font=\scriptsize,
    inner xsep=2pt,
    inner ysep=4pt,
]
\begin{figure*}[!t]
    \centering
    \resizebox{\textwidth}{!}{
        \begin{forest}
            forked edges,
            for tree={
                grow=east,
                reversed=true,
                anchor=base west,
                parent anchor=east,
                child anchor=west,
                base=left,
                font=\sffamily\small,
                rectangle,
                line width=0.8pt,
                draw=darkblue,
                fill=lightblue!60,
                rounded corners,
                align=left,
                minimum width=3em,
                edge+={darkblue, line width=0.8pt},
                s sep=3pt,
                inner xsep=2pt,
                inner ysep=3pt,
                ver/.style={rotate=90, child anchor=north, parent anchor=south, anchor=center},
            },
            where level=1{text width=3.7em,font=\sffamily\scriptsize,edge+={rounded corners}}{},
            where level=2{text width=6.3em,font=\scriptsize,}{},
            where level=3{text width=7.0em,font=\scriptsize,}{},
            where level=4{text width=6.1em,font=\scriptsize,}{},
            [
                \textbf{MLLM Evaluation}, ver, l sep=1pt, draw=rootdarkblue,fill=rootlightblue,line width=1.4pt,font=\sffamily,
                    [
                        Keyword \\ Extraction,draw=darkred,fill=lightred,edge+={darkred, line width=0.8pt},
                        [
                                TRI-HE~\cite{trihe}{, }
                                MERLIM~\cite{merlim}{, }
                                AMBER~\cite{amber}{, }
                                NLTK~\cite{nltk}{, }
                                WordNet~\cite{wordnet}{, }
                                SpaCy~\cite{honnibal2020spacy}{, }
                                FaithScore~\cite{jing2023faithscore}{, }
                                SSGP~\cite{ssgp}{, }\\
                                FCGP~\cite{pcfg}{, }
                                SPICE~\cite{spice}{, }
                                RAGChecker~\cite{ragchecker}{, }
                                HaloQuest~\cite{haloquest}{, }
                                VALOR~\cite{valor}{, }
                                MOCHA~\cite{mocha}{, }
                                MMArXiv~\cite{mmarxiv}{, }
                                , leaf, text width=40.5em,draw=darkred,fill=lightred,edge+={darkred, line width=0.8pt},
                        ]
                    ]
                    [
                        Embedding \\Based \\Evaluation,draw=darkgreen,fill=lightgreen,edge+={darkgreen, line width=0.8pt},
                        [
                                TRI-HE~\cite{trihe}{, }
                                RAGAS~\cite{ragas}{, }
                                FIHA~\cite{fiha}{, }
                                Pfram~\cite{pfram}{, }
                                Reefknot~\cite{reefknot}{, }
                                DBD~\cite{dbd}{, }
                                MediHallDetector~\cite{medihalldetector}{, }
                                CLIP~\cite{clip}{, }\\
                                MetaToken~\cite{metatoken}{, }
                                MOCHA~\cite{mocha}
                                , leaf, text width=40.5em,draw=darkgreen,fill=lightgreen,edge+={darkgreen, line width=0.8pt},
                        ]
                    ]
                    [
                       MLLM \\Based \\ Evaluation,draw=darkyellow,fill=lightyellow,edge+={darkyellow, line width=0.8pt}, 
                        [
                               LLaVA-Bench~\cite{liu2023visual}{, }
                               TRI-HE~\cite{trihe}{, }
                               AVHBench~\cite{avhbench}{, }
                               GAVIE~\cite{lrv}{, }
                               RAGAS~\cite{ragas}{, }
                               LongHalQA~\cite{longhalqa}{, }
                               VCD~\cite{vcd}{, }\\
                               VALOR~\cite{valor}{, }
                               HaloQuest~\cite{haloquest}{, }
                               Hal-Eval~\cite{haleval}{, }
                               MOCHA~\cite{mocha}{, }
                               HallusionBench~\cite{hallusionbench}{, }
                               AGLA~\cite{agla}{, }
                               LLaVA-RLHF~\cite{llavarlhf}
                                , leaf, text width=40.5em,draw=darkyellow,fill=lightyellow,edge+={darkyellow, line width=0.8pt},
                        ]
                    ]
                    [
                        Evaluation \\Platform,
                        [
                            VLMEvalKit~\cite{vlmevalkit}{, }
                            WildVision~\cite{wildvision}{, }
                            LMMs-Eval~\cite{lmmseval}{, }
                            MultiMedEval~\cite{multimedeval}{, }
                            AgentStudio~\cite{agentstudio}{, }
                            Video-Bench~\cite{videobench}
                            , leaf, text width=40.5em,
                        ]
                    ]
            ]
        \end{forest}
    }
    \caption{Categories of MLLM benchmarks. }
    \label{taxo_of_eval}
\end{figure*}
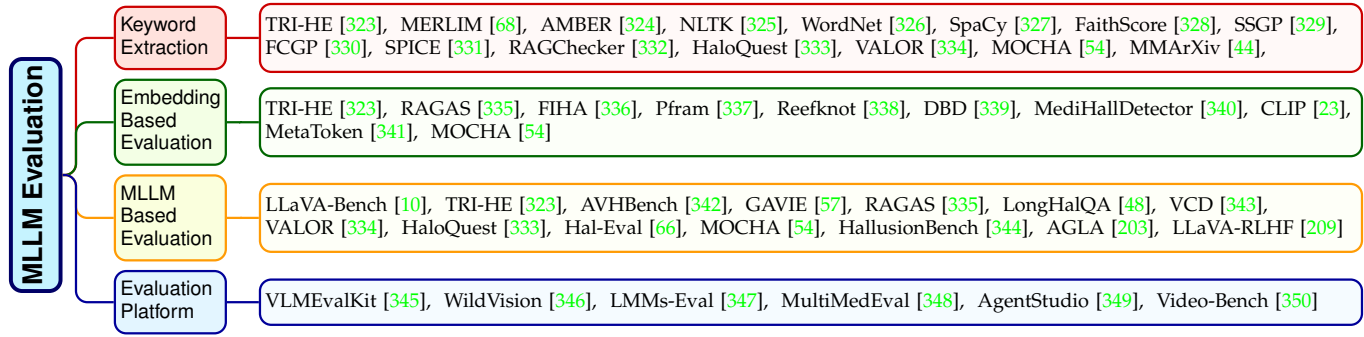

\subsection{Video Perception}
Building on the remarkable success of MLLMs in image perception~\cite{llavanext,liu2023visual}, recent studies~\cite{ren2024timechat,zhang2023video,lin2023video} have explored their potential in video perception tasks. In this section, we review how external tools augment MLLMs across three representative video tasks: Temporal Grounding, Dense Captioning, and Question Answering.

\subsubsection{Temporal Grounding}
Temporal Grounding aims to identify the start and end times of an event described by natural language within a given video~\cite{gao2017tall}. The primary challenge is to comprehend the entire video context and output precise numerical event timestamps~\cite{liu2024bench}. Several studies explore leveraging external tools to address this challenging task. For example, TFVTG~\cite{zheng2024training} leverages similarity scores derived from BLIP-2~\cite{li2022blip} between the descriptions of each proposed timeslot and the query event to filter and integrate these proposals, thereby generating the final prediction. ChatVTG~\cite{qu2024chatvtg} employs SentenceBERT~\cite{sentencebert} to encode the captions of each timeslot together with the query event, subsequently selecting the timeslot with the highest similarity score as the prediction. Momentor~\cite{qian2024momentor} utilizes specialized temporal encoders and decoders for precise time localization. VTG-LLM~\cite{guo2025vtg} and E.T.Chat~\cite{liu2024bench} employ token compressors~\cite{li2022blip} to increase the number of sampled frames, thereby enhancing the precision of grounding. TRACE~\cite{guo2024trace} introduces a specialized temporal decoder to generate numerical time representations. VideoMind~\cite{videomind} utilizes a grounded agent to handle grounding-related tasks. Time-R1~\cite{wang2025time} and VideoChat-R1~\cite{li2025videochat} introduce a verifiable reward model to enhance temporal localization capabilities through a reinforcement learning framework. Furthermore, MUSEG~\cite{luo2025museg} extends this reward model into a multi-grained paradigm.

\subsubsection{Dense Captioning}
Dense Captioning aims to generate descriptive captions for multiple events within a video while simultaneously localizing their temporal boundaries~\cite{krishna2017dense}. The challenges of this task stem from the need for both temporal grounding and fine-grained perception across the entire video, thereby necessitating the integration of external tools. 
For example, VTG-LLM~\cite{guo2025vtg} and E.T.Chat~\cite{liu2024bench} employ token compressors~\cite{li2022blip} to increase the number of sampled frames, thereby enhancing the richness of the generated descriptions. TRACE~\cite{guo2024trace} introduces a causal encoding architecture combined with a specialized text encoder to incorporate all preceding content for producing precise captions. VideoMind~\cite{videomind} leverages an answering agent to process outputs from other agents and generate the final captions. Time-R1~\cite{wang2025time}, VideoChat-R1~\cite{li2025videochat}, and VersaVid~\cite{chen2025versavid} incorporate a verifiable reward model to guide the generation of appropriate answer templates through a reinforcement learning framework. Furthermore, MUSEG~\cite{luo2025museg} and VideoCap-R1~\cite{meng2025videocap} extend this reward model into a multi-grained and multi-step paradigm. Lastly, PAM~\cite{lin2025perceive} generates captions by employing a prompt encoder to utilize externally provided prompts.

\subsubsection{Question Answering}
Different from temporal grounding and dense captioning, Question Answering evaluates the video perception capabilities of MLLMs from a broader and more diverse perspective~\cite{li2024mvbench,fu2024video}. Therefore, many studies have explored leveraging external tools for general tasks or specific downstream requirements.  For example, MoReVQA~\cite{min2024morevqa} presents a multi-stage system composed of several tailored LLMs to address challenging questions. VersaVid~\cite{chen2025versavid} introduces a task-specific reward model to enhance the reasoning capabilities for diverse questions. DeepVideo-R1~\cite{park2025deepvideo} redesigns the reward model in a regression-based manner. VideoGPT+~\cite{maaz2024videogpt+} incorporates an image encoder~\cite{clip} to extract more detailed spatial information. TraveLER~\cite{shang2024traveler}, VideoAgent~\cite{videoagent}, and MUPA~\cite{dang2025mupa} leverage a series of agents to decompose questions and answer them using specialized tools. PAM~\cite{lin2025perceive} tackles question answering by employing a prompt encoder to incorporate externally provided prompts. LangRepo~\cite{kahatapitiya2024language} introduces a language repository to maintain concise and structured information as an interpretable representation within long-form videos. HierarQ~\cite{azad2025hierarq} utilizes multiple Q-formers~\cite{li2022blip} to effectively capture both short and long-term context. MVU~\cite{ranasinghe2024understanding} uses a vision detector~\cite{minderer2022simple} to extract object-centric information. LLoVi~\cite{zhang2023simple} uses an external video captioner~\cite{zhao2023learning} to generate short-term captions, which assist the understanding of long videos. LVNet~\cite{park2024too} leverages a key frame selector to focus on the most relevant content within long videos. OmAgent~\cite{zhang2024omagent} incorporates a set of agents for processing long videos. DrVideo~\cite{drvideo} transforms a long video into a document and then answers the questions with textual guidance.
\section{Evaluation}

\label{evaluation}
Given the complexity of multimodal data and the open-ended nature of generative tasks, developing robust evaluation metrics that accurately measure answer quality and text-visual consistency remains a grand challenge. It restricts our ability to reliably assess model performance and, in turn, impedes the development of more generalizable and trustworthy MLLMs. To this end, external tools have been increasingly adopted to support more comprehensive and precise evaluation. This section presents four representative approaches that employ external tools for MLLM evaluations: \textit{Keyword Extraction}, \textit{Embedding-based Evaluation}, \textit{MLLM-based Evaluation}, and \textit{Evaluation Platform}.

\subsection{Keyword Extraction}
Unlike traditional discriminative tasks with fixed answer candidates, MLLM-related tasks often involve open-ended generation, where responses may include extraneous or irrelevant content. Extracting task-relevant keywords from the MLLM-generated responses has therefore been widely adopted for accurate evaluations in downstream tasks.

To identify task-relevant keywords, many studies leverage external libraries and parsers.
MERLIM~\cite{merlim} uses the SpaCy library~\cite{honnibal2020spacy} to check words in responses and determine if it is a nouns or not. Then the extracted nouns are compared with ground-truth objects to evaluate model performance. 
To better manage the extensive vocabulary output of MLLMs, the authors also utilize the WordNet hierarchy~\cite{wordnet} and ChatGPT~\cite{chatgpt} to identify potential synonyms for the extracted nouns.
AMBER~\cite{amber} first extracts nouns from responses by language toolkits, NLTK~\cite{nltk}, to obtain the initial objects, and filters unnecessary objects that are not in the pre-defined list to obtain the final objects for evaluation.
SPICE~\cite{spice} employs a Stanford scene graph parser~\cite{ssgp} and a probabilistic context-free grammar dependency parser~\cite{pcfg} to construct scene graphs based on the responses, which is frequently adopted to assess the quality of generated captions.
RAGChecker~\cite{ragchecker} utilizes a text-to-claim extractor to decompose a given response into a set of claims, which provides a more fine-grained evaluation for the multimodal retrieval augmented system.

Beyond external libraries, powerful large models, such as GPT-4V~\cite{gpt4o}, are also widely adopted. 
For example, TRI-HE~\cite{trihe} prompts GPT-4V~\cite{gpt4o} to extract object and relation triplets from the responses, and the extracted triplets are used to evaluate model performance on object and relation understanding.
FaithScore~\cite{jing2023faithscore} employs ChatGPT~\cite{chatgpt} as a recognizer to identify descriptive sub-sentences from the responses that require further verification.
VALOR~\cite{valor} employs GPT-4V~\cite{gpt4o} to extract pivotal features that are extracted from the generated descriptions for subsequent evaluation.
HaloQuest~\cite{haloquest} employs a Langfun schema with Gemini~\cite{gemini} to accurately extract the main point in the model response for performance evaluation.
MOCHA~\cite{mocha} performs object parsing with ChatGPT~\cite{chatgpt} to extract objects in the responses and compare them with ground-truth objects through LLMs and Natural Language Inference models.
MMArXiv~\cite{mmarxiv} designs a prompt to employ GPT-4V~\cite{gpt4o} for extracting the answer label from the model generations.

\subsection{Embedding-based Evaluation}
Embedding-based evaluation is usually performed by external models that can measure similarities between two samples. For example, CLIP~\cite{clip} can be utilized to measure the consistency between the image and responses, and the Natural Language Inference (NLI) model~\cite{sentencebert} can be utilized to evaluate the matching degree between the responses and ground-truth answers.

For instance, TRI-HE~\cite{trihe} employs Sentence-BERT~\cite{sentencebert} to calculate cosine similarity between the answer triplets and all triplets in the image scene graph, and the triplets with similarities greater than a threshold are considered as true responses.
While Pfram~\cite{pfram}, FIHA~\cite{fiha}, MOCHA~\cite{mocha}, and Reefknot~\cite{reefknot} directly employ embedding-based BERT models~\cite{sentencebert} to calculate cosine similarity between ground-truth descriptions and the model outputs for generation ability evaluation.
RAGAS~\cite{ragas} employs the text-embedding-ada-002 API to calculate the similarity between the original question and the questions generated based on the LLM responses, which can evaluate how closely the generated answer aligns with the initial question or instruction.
MediHallDetector~\cite{medihalldetector} and MetaToken~\cite{metatoken} employ CLIP~\cite{clip} to calculate similarity between the generated captions and the input images, while DBD~\cite{dbd} improves this schema with a novel set of CLIP-based metrics to better reflect the quality of captions.

\subsection{MLLM-based Evaluation}
Leveraging the advanced instruction-following capabilities of powerful MLLMs, such as GPT-4V~\cite{gpt4o} and Gemini~\cite{gemini}, it is natural to leverage these MLLMs to evaluate the overall quality of model-generated outputs. 

For example, LLaVA-Bench~\cite{liu2023visual} utilizes detailed prompts to instruct GPT-4~\cite{gpt4o} to evaluate the helpfulness, relevance, accuracy, and level of detail of the responses, and then provide an overall score on a scale from 1 to 10 with additional reasons as final judgment.
TRI-HE~\cite{trihe} leverages prompting of a powerful LLM, GPT-4V~\cite{gpt4o}, to determine whether a given extracted triplet can be directly obtained or inferred from an image scene graph.
AVHBench~\cite{avhbench} employs the GAVIE~\cite{lrv} metric with GPT-4~\cite{gpt4o} to measure the accuracy of the generated captions.
RAGAS~\cite{ragas} employs ChatGPT~\cite{chatgpt} to measure the faithfulness of the generated responses and prompt it to generate potential questions based on the responses.
LongHalQA~\cite{longhalqa} utilizes GPT-4~\cite{gpt4o} to generate data and evaluate MLLM performance on long and complex hallucination text.
HaloQuest~\cite{haloquest} employs a Langfun schema to help Gemini~\cite{gemini} accurately extract the main point in the model response and ground truth, and then decide whether these points are in agreement.
VALOR~\cite{valor}, VCD~\cite{vcd}, and HallusionBench~\cite{hallusionbench} employ GPT-4V~\cite{gpt4o} as the evaluation LLM agent to evaluate faithfulness and convergence of the generated captions.
Hal-Eval~\cite{haleval} utilizes a fine-tuned LLaVA~\cite{liu2023visual} as a reference-free, open-source evaluation model to evaluate MLLM performance on multimodal hallucination.

\subsection{Evaluation Platform}
With the rapid advancement of MLLM backbones, evaluation benchmarks, and metrics, assessing the capabilities of MLLMs across diverse aspects has become increasingly challenging and time-consuming. Numerous MLLM evaluation platforms have therefore been developed, integrating multiple models, datasets, and evaluation metrics to address the evaluation challenge. These platforms enable more effective and efficient evaluation of MLLMs and accelerate progress in the field of multimodal learning.

For instance, VLMEvalKit~\cite{vlmevalkit} is an open-source, user-friendly, and comprehensive toolkit for evaluating MLLMs, which covers over 200+ MLLMs and 80+ multimodal benchmarks. The toolkit can automatically handle the entire evaluation process, including data preparation, distributed inference, prediction, post-processing, and metric calculation.
% LVLM-eHub~\cite{lvlmehub} comprises 8 representative MLLMs and 47 standard benchmarks across 6 categories for evaluating multimodal capabilities. In addition to offering a comprehensive evaluation platform, the authors present several insightful findings for future research.
LMMs-Eval~\cite{lmmseval} is a unified and standardized multimodal benchmark framework with over 50 tasks and 10 models to promote transparent and reproducible MLLM performance evaluations. 

For domain-specific MLLM evaluation, MultiMedEval~\cite{multimedeval} is introduced to evaluate medical MLLMs, encompassing 6 medical tasks and 23 datasets. 
WildVision~\cite{wildvision} is an online platform that collects human preferences to evaluate MLLMs. By selecting 500 high-quality samples and employing GPT-4~\cite{gpt4o} as the judge, the proposed benchmark achieves a Spearman correlation of 0.94 that significantly outperforms other benchmarks.
AgentStudio~\cite{agentstudio} is a trinity of environments, tools, and benchmarks to evaluate MLLM on agent-related tasks, providing a lightweight, interactive environment with highly generic observation and action spaces for creating tasks, labeling actions, and automatic evaluations. 
Video-Bench~\cite{videobench} is a new comprehensive benchmark along with a toolkit specifically designed for evaluating Video-LLMs, facilitating the calculation of metrics and generating convenient final scores. 
\section{Challenges and Future}
\label{future}
% Efficiency
% Privacy
% Proactive
% Reliability and Robustness
% Cost
% Fairness (evaluation)
% More modalities
Despite the enhanced knowledge and performance that external tools bring to MLLMs, several critical challenges remain unresolved and underexplored. These challenges may represent key bottlenecks in the pursuit of human-level multimodal intelligence.
In the following, we summarize the key challenges and potential research directions for the future development of external tool-empowered MLLMs.
\begin{itemize}
    \item \textbf{Efficiency.} Incorporating external tools into downstream multimodal tasks can introduce additional inference latency, potentially hindering efficient deployment of MLLMs.
    Therefore, a promising research direction is to explore how external tools could be integrated during the training phase, enabling MLLMs to internalize expert capabilities, thereby reducing reliance on tools at inference time and mitigating efficiency issues.
    \item  \textbf{Privacy.} Using API-based models for data annotation or performance evaluation may pose serious concerns in data privacy and safety. Therefore, masking sensitive information in the data or developing open-source models that can be deployed locally are promising strategies for domains where data privacy is critical.
    \item \textbf{Proactivity.} The prevailing approach of using external tools is manually decided by humans, regardless of whether to use external tools or which ones to use. MLLMs lack both the initiative to invoke tools autonomously and the ability to make informed decisions about which tools to use.
    Therefore, developing agent-based MLLMs that can autonomously decide when and which tools to use is a crucial step toward achieving artificial general intelligence.
    \item \textbf{Trustworthiness.} The performance of tool-enhanced MLLMs depends heavily on the quality and reliability of the external models they rely on, which may not be trustworthy across challenging tasks.
    As a result, a promising research direction is to allow them to adaptively determine when and how to trust external outputs and thereby improve the robustness and trustworthiness of tool-empowered MLLMs.
    \item \textbf{Cost.} Using expert models to empower MLLMs can incur significant API costs or consume external GPU resources. Therefore, a key research direction is to explore cost-efficient integration strategies, such as compressing tool functionalities into the MLLM itself, caching tool outputs, or employing lightweight alternatives, to mitigate dependency on costly external resources.
    \item \textbf{Fairness.} Using external large models for performance evaluation may raise concerns about comparison fairness, as these models can exhibit clear biases toward specific output styles or content, especially when ground-truth answers are not available.
    To mitigate this issue, it is beneficial to incorporate ground-truth answers into the evaluation process, providing a more consistent and reliable reference for comparison.
    \item  \textbf{Diversity.} The types of tools and the ways they are utilized in current MLLMs remain limited. Therefore, a promising research direction is to incorporate a broader range of tools and explore more diverse and flexible strategies for their integration, which may help advance tool-empowered MLLMs to the next level.
\end{itemize}
\section{Conclusion}
In conclusion, this survey highlights the significant potential of integrating external tools to address the existing limitations of Multimodal Large Language Models. By enhancing data quality, boosting performance on complex tasks, and enabling more rigorous evaluation, tool-augmented MLLMs represent a crucial step toward more reliable and versatile multimodal intelligence. Our comprehensive analysis not only reveals current challenges but also outlines promising future directions, emphasizing that the strategic use of external resources will be key to unlocking the next generation of MLLM capabilities and advancing the pursuit of artificial general intelligence.
%\clearpage

{
\bibliographystyle{revision_ref}
\bibliography{references}

\begin{thebibliography}{100}

\bibitem{chatgpt}
J.~Achiam et~al.
\newblock Gpt-4 technical report.
\newblock {\em arXiv:2303.08774}, 2023.

\bibitem{touvron2023llama}
H.~Touvron et~al.
\newblock Llama: Open and efficient foundation language models.
\newblock {\em arXiv:2302.13971}, 2023.

\bibitem{vicuna2023}
W.-L. Chiang et~al.
\newblock Vicuna: An open-source chatbot impressing gpt-4 with 90\%* chatgpt quality, 2023.

\bibitem{yang2025qwen3}
A.~Yang et~al.
\newblock Qwen3 technical report.
\newblock {\em arXiv:2505.09388}, 2025.

\bibitem{liu2024deepseek}
A.~Liu et~al.
\newblock Deepseek-v3 technical report.
\newblock {\em arXiv:2412.19437}, 2024.

\bibitem{loop}
W.~An et~al.
\newblock Generalized category discovery with large language models in the loop.
\newblock In {\em ACL Findings}, 2024.

\bibitem{kasneci2023chatgpt}
E.~Kasneci et~al.
\newblock Chatgpt for good? on opportunities and challenges of large language models for education.
\newblock {\em Learning and individual differences}, 2023.

\bibitem{dka}
W.~An et~al.
\newblock Knowledge acquisition disentanglement for knowledge-based visual question answering with large language models.
\newblock {\em arXiv:2407.15346}, 2024.

\bibitem{dai2023instructblip}
W.~Dai et~al.
\newblock Instructblip: Towards general-purpose vision-language models with instruction tuning.
\newblock {\em arXiv:2306.04387}, 2023.

\bibitem{liu2023visual}
H.~Liu et~al.
\newblock Visual instruction tuning.
\newblock {\em arXiv:2304.08485}, 2023.

\bibitem{zhu2025internvl3}
J.~Zhu et~al.
\newblock Internvl3: Exploring advanced training and test-time recipes for open-source multimodal models.
\newblock {\em arXiv:2504.10479}, 2025.

\bibitem{bai2025qwen25vl}
S.~Bai et~al.
\newblock Qwen2. 5-vl technical report.
\newblock {\em arXiv:2502.13923}, 2025.

\bibitem{wu2024deepseek}
Z.~Wu et~al.
\newblock Deepseek-vl2: Mixture-of-experts vision-language models for advanced multimodal understanding.
\newblock {\em arXiv:2412.10302}, 2024.

\bibitem{mllmsurvey}
S.~Yin et~al.
\newblock A survey on multimodal large language models.
\newblock {\em National Science Review}, November 2024.

\bibitem{lin2024schedule}
H.~Lin et~al.
\newblock Schedule your edit: A simple yet effective diffusion noise schedule for image editing.
\newblock {\em NeurIPS}, 2024.

\bibitem{surveyhallucination}
Z.~Bai et~al.
\newblock Hallucination of multimodal large language models: A survey.
\newblock {\em arXiv:2404.18930}, 2024.

\bibitem{zhang2024redundancy}
X.~Zhang et~al.
\newblock From redundancy to relevance: Enhancing explainability in multimodal large language models.
\newblock {\em arXiv e-prints}, pp. arXiv--2406, 2024.

\bibitem{sun2024review}
S.~Sun et~al.
\newblock A review of multimodal explainable artificial intelligence: Past, present and future.
\newblock {\em arXiv preprint arXiv:2412.14056}, 2024.

\bibitem{surveyreasoning}
Y.~Wang et~al.
\newblock Multimodal chain-of-thought reasoning: A comprehensive survey.
\newblock {\em arXiv:2503.12605}, 2025.

\bibitem{hu2023advancing}
M.~Hu et~al.
\newblock Advancing medical imaging with language models: A journey from n-grams to chatgpt.
\newblock {\em arXiv:2304.04920}, 2023.

\bibitem{chen2023driving}
L.~Chen et~al.
\newblock Driving with llms: Fusing object-level vector modality for explainable autonomous driving.
\newblock {\em arXiv:2310.01957}, 2023.

\bibitem{gpt4o}
A.~Hurst et~al.
\newblock Gpt-4o system card.
\newblock {\em arXiv:2410.21276}, 2024.

\bibitem{clip}
A.~Radford et~al.
\newblock Learning transferable visual models from natural language supervision.
\newblock In {\em ICML}, 2021.

\bibitem{liu2021swin}
Z.~Liu et~al.
\newblock Swin transformer: Hierarchical vision transformer using shifted windows.
\newblock In {\em ICCV}, 2021.

\bibitem{elizalde2023clap}
B.~Elizalde et~al.
\newblock Clap learning audio concepts from natural language supervision.
\newblock In {\em ICASSP}, 2023.

\bibitem{laurenccon2024matters}
H.~Lauren{\c{c}}on et~al.
\newblock What matters when building vision-language models?
\newblock {\em arXiv:2405.02246}, 2024.

\bibitem{bai2023qwenllm}
J.~Bai et~al.
\newblock Qwen technical report.
\newblock {\em arXiv:2309.16609}, 2023.

\bibitem{qin2024tool}
Y.~Qin et~al.
\newblock Tool learning with foundation models.
\newblock {\em ACM Computing Surveys}, 2024.

\bibitem{gemini}
G.~Team et~al.
\newblock Gemini 1.5: Unlocking multimodal understanding across millions of tokens of context.
\newblock {\em arXiv:2403.05530}, 2024.

\bibitem{vlhtest}
W.~Huang et~al.
\newblock Visual hallucinations of multi-modal large language models.
\newblock {\em arXiv:2402.14683}, 2024.

\bibitem{correlationqa}
T.~Han et~al.
\newblock The instinctive bias: Spurious images lead to illusion in mllms.
\newblock {\em arXiv:2402.03757}, 2024.

\bibitem{mmvp}
S.~Tong et~al.
\newblock Eyes wide shut? exploring the visual shortcomings of multimodal llms.
\newblock In {\em CVPR}, 2024.

\bibitem{concept}
P.~Sharma et~al.
\newblock Conceptual captions: A cleaned, hypernymed, image alt-text dataset for automatic image captioning.
\newblock In {\em ACL}, 2018.

\bibitem{flume}
C.~Chambers et~al.
\newblock Flumejava: easy, efficient data-parallel pipelines.
\newblock {\em ACM Sigplan Notices}, 2010.

\bibitem{pali}
X.~Chen et~al.
\newblock Pali: A jointly-scaled multilingual language-image model.
\newblock {\em arXiv:2209.06794}, 2022.

\bibitem{laion}
C.~Schuhmann et~al.
\newblock Laion-5b: An open large-scale dataset for training next generation image-text models.
\newblock {\em NeurIPS}, 2022.

\bibitem{datacomp}
S.~Y. Gadre et~al.
\newblock Datacomp: In search of the next generation of multimodal datasets.
\newblock {\em NeurIPS}, 2023.

\bibitem{align}
C.~Jia et~al.
\newblock Scaling up visual and vision-language representation learning with noisy text supervision.
\newblock In {\em ICML}, 2021.

\bibitem{mmc4}
W.~Zhu et~al.
\newblock Multimodal c4: An open, billion-scale corpus of images interleaved with text.
\newblock {\em NeurIPS}, 2023.

\bibitem{wukong}
J.~Gu et~al.
\newblock Wukong: A 100 million large-scale chinese cross-modal pre-training benchmark.
\newblock {\em NeurIPS}, 2022.

\bibitem{wit}
K.~Srinivasan et~al.
\newblock Wit: Wikipedia-based image text dataset for multimodal multilingual machine learning.
\newblock In {\em ACM SIGIR conference on research and development in information retrieval}, 2021.

\bibitem{conceptual12m}
S.~Changpinyo et~al.
\newblock Conceptual 12m: Pushing web-scale image-text pre-training to recognize long-tail visual concepts.
\newblock In {\em CVPR}, 2021.

\bibitem{redcaps}
K.~Desai et~al.
\newblock Redcaps: Web-curated image-text data created by the people, for the people.
\newblock {\em arXiv:2111.11431}, 2021.

\bibitem{mmarxiv}
L.~Li et~al.
\newblock Multimodal arxiv: A dataset for improving scientific comprehension of large vision-language models.
\newblock {\em arXiv:2403.00231}, 2024.

\bibitem{vqav2}
Y.~Goyal et~al.
\newblock Making the v in vqa matter: Elevating the role of image understanding in visual question answering.
\newblock In {\em CVPR}, 2017.

\bibitem{aic}
J.~Wu et~al.
\newblock Ai challenger: A large-scale dataset for going deeper in image understanding.
\newblock {\em arXiv:1711.06475}, 2017.

\bibitem{im2text}
V.~Ordonez et~al.
\newblock Im2text: Describing images using 1 million captioned photographs.
\newblock {\em NeurIPS}, 2011.

\bibitem{longhalqa}
H.~Qiu et~al.
\newblock Longhalqa: Long-context hallucination evaluation for multimodal large language models.
\newblock {\em arXiv:2410.09962}, 2024.

\bibitem{autohallusion}
X.~Wu et~al.
\newblock Autohallusion: Automatic generation of hallucination benchmarks for vision-language models.
\newblock {\em arXiv:2406.10900}, 2024.

\bibitem{throne}
P.~Kaul et~al.
\newblock Throne: An object-based hallucination benchmark for the free-form generations of large vision-language models.
\newblock In {\em CVPR}, 2024.

\bibitem{phd}
J.~Liu et~al.
\newblock Phd: A chatgpt-prompted visual hallucination evaluation dataset.
\newblock {\em arXiv:2403.11116}, 2024.

\bibitem{aigcs}
Y.~Gao et~al.
\newblock Aigcs confuse ai too: Investigating and explaining synthetic image-induced hallucinations in large vision-language models.
\newblock In {\em ACMMM}, 2024.

\bibitem{easy}
Y.~Qian et~al.
\newblock How easy is it to fool your multimodal llms? an empirical analysis on deceptive prompts.
\newblock {\em arXiv:2402.13220}, 2024.

\bibitem{mocha}
A.~Ben-Kish et~al.
\newblock Mitigating open-vocabulary caption hallucinations.
\newblock {\em arXiv:2312.03631}, 2023.

\bibitem{fohe}
L.~Wang et~al.
\newblock Mitigating fine-grained hallucination by fine-tuning large vision-language models with caption rewrites.
\newblock In {\em International Conference on Multimedia Modeling}, 2024.

\bibitem{nope}
H.~Lovenia et~al.
\newblock Negative object presence evaluation (nope) to measure object hallucination in vision-language models.
\newblock {\em arXiv:2310.05338}, 2023.

\bibitem{lrv}
F.~Liu et~al.
\newblock Mitigating hallucination in large multi-modal models via robust instruction tuning.
\newblock {\em arXiv:2306.14565}, 2023.

\bibitem{ciem}
H.~Hu et~al.
\newblock Ciem: Contrastive instruction evaluation method for better instruction tuning.
\newblock {\em arXiv:2309.02301}, 2023.

\bibitem{mmhalsnowball}
W.~Zhong et~al.
\newblock Investigating and mitigating the multimodal hallucination snowballing in large vision-language models.
\newblock {\em arXiv:2407.00569}, 2024.

\bibitem{silkie}
L.~Li et~al.
\newblock Silkie: Preference distillation for large visual language models.
\newblock {\em arXiv:2312.10665}, 2023.

\bibitem{dvqa}
K.~Kafle et~al.
\newblock Dvqa: Understanding data visualizations via question answering.
\newblock In {\em CVPR}, 2018.

\bibitem{mmrel}
J.~Nie et~al.
\newblock Mmrel: A relation understanding dataset and benchmark in the mllm era.
\newblock {\em arXiv e-prints}, pp. arXiv--2406, 2024.

\bibitem{invest}
Y.~Liu et~al.
\newblock Investigating and mitigating object hallucinations in pretrained vision-language (clip) models.
\newblock {\em arXiv:2410.03176}, 2024.

\bibitem{vga}
M.~Ziyang et~al.
\newblock Vga: Vision gui assistant-minimizing hallucinations through image-centric fine-tuning.
\newblock In {\em EMNLP Findings}, 2024.

\bibitem{aloha}
S.~Petryk et~al.
\newblock Aloha: A new measure for hallucination in captioning models.
\newblock {\em arXiv:2404.02904}, 2024.

\bibitem{haleval}
C.~Jiang et~al.
\newblock Hal-eval: A universal and fine-grained hallucination evaluation framework for large vision language models.
\newblock In {\em ACMMM}, 2024.

\bibitem{easydetect}
X.~Chen et~al.
\newblock Unified hallucination detection for multimodal large language models.
\newblock {\em arXiv:2402.03190}, 2024.

\bibitem{merlim}
A.~Villa et~al.
\newblock Behind the magic, merlim: Multi-modal evaluation benchmark for large image-language models.
\newblock In {\em CVPR}, 2025.

\bibitem{rah}
Z.~Chen et~al.
\newblock Mitigating hallucination in visual language models with visual supervision.
\newblock {\em arXiv:2311.16479}, 2023.

\bibitem{hallecontrol}
B.~Zhai et~al.
\newblock Halle-control: controlling object hallucination in large multimodal models.
\newblock {\em arXiv:2310.01779}, 2023.

\bibitem{haelm}
J.~Wang et~al.
\newblock Evaluation and analysis of hallucination in large vision-language models.
\newblock {\em arXiv:2308.15126}, 2023.

\bibitem{pope}
Y.~Li et~al.
\newblock Evaluating object hallucination in large vision-language models.
\newblock {\em arXiv:2305.10355}, 2023.

\bibitem{som}
A.~Yan et~al.
\newblock List items one by one: A new data source and learning paradigm for multimodal llms.
\newblock {\em arXiv:2404.16375}, 2024.

\bibitem{mhaldetect}
A.~Gunjal et~al.
\newblock Detecting and preventing hallucinations in large vision language models.
\newblock In {\em AAAI}, 2024.

\bibitem{hallucinogen}
A.~Seth et~al.
\newblock Towards a systematic evaluation of hallucinations in large-vision language models.
\newblock {\em arXiv:2412.20622}, 2024.

\bibitem{videohallucer}
Y.~Wang et~al.
\newblock Videohallucer: Evaluating intrinsic and extrinsic hallucinations in large video-language models.
\newblock {\em arXiv:2406.16338}, 2024.

\bibitem{illusionvqa}
H.~S. Shahgir et~al.
\newblock Illusionvqa: A challenging optical illusion dataset for vision language models.
\newblock {\em arXiv:2403.15952}, 2024.

\bibitem{vdgd}
S.~Ghosh et~al.
\newblock Visual description grounding reduces hallucinations and boosts reasoning in lvlms.
\newblock {\em arXiv:2405.15683}, 2024.

\bibitem{vidhalluc}
C.~Li et~al.
\newblock Vidhalluc: Evaluating temporal hallucinations in multimodal large language models for video understanding.
\newblock In {\em CVPR}, 2025.

\bibitem{coyo}
M.~Byeon et~al.
\newblock Coyo-700m: Image-text pair dataset, 2022.

\bibitem{obelics}
H.~Lauren{\c{c}}on et~al.
\newblock Obelics: An open web-scale filtered dataset of interleaved image-text documents.
\newblock {\em NeurIPS}, 2023.

\bibitem{oquab2023dinov2}
M.~Oquab et~al.
\newblock Dinov2: Learning robust visual features without supervision.
\newblock {\em arXiv:2304.07193}, 2023.

\bibitem{stablediffusion}
D.~Podell et~al.
\newblock Sdxl: Improving latent diffusion models for high-resolution image synthesis.
\newblock {\em arXiv:2307.01952}, 2023.

\bibitem{dalle}
J.~Betker et~al.
\newblock Improving image generation with better captions.
\newblock {\em Computer Science}, 2023.

\bibitem{seem}
X.~Zou et~al.
\newblock Segment everything everywhere all at once.
\newblock {\em NeurIPS}, 2023.

\bibitem{groundingdino}
S.~Liu et~al.
\newblock Grounding dino: Marrying dino with grounded pre-training for open-set object detection.
\newblock In {\em ECCV}, 2024.

\bibitem{ram}
X.~Huang et~al.
\newblock Open-set image tagging with multi-grained text supervision.
\newblock {\em arXiv:2310.15200}, 2023.

\bibitem{li2022blip}
J.~Li et~al.
\newblock Blip: Bootstrapping language-image pre-training for unified vision-language understanding and generation.
\newblock In {\em ICML}, 2022.

\bibitem{gme}
X.~Zhang et~al.
\newblock Gme: Improving universal multimodal retrieval by multimodal llms.
\newblock {\em arXiv:2412.16855}, 2024.

\bibitem{contriever}
G.~Izacard et~al.
\newblock Unsupervised dense information retrieval with contrastive learning.
\newblock {\em arXiv:2112.09118}, 2021.

\bibitem{pic2word}
K.~Saito et~al.
\newblock Pic2word: Mapping pictures to words for zero-shot composed image retrieval.
\newblock In {\em CVPR}, 2023.

\bibitem{xlheadtags}
F.~T. Shohan et~al.
\newblock Xl-headtags: Leveraging multimodal retrieval augmentation for the multilingual generation of news headlines and tags.
\newblock {\em arXiv:2406.03776}, 2024.

\bibitem{echosight}
Y.~Yan and W.~Xie.
\newblock Echosight: Advancing visual-language models with wiki knowledge.
\newblock {\em arXiv:2407.12735}, 2024.

\bibitem{eclip}
Y.~Kumar and P.~Marttinen.
\newblock Improving medical multi-modal contrastive learning with expert annotations.
\newblock In {\em ECCV}, 2024.

\bibitem{visa}
X.~Ma et~al.
\newblock Visa: Retrieval augmented generation with visual source attribution.
\newblock {\em arXiv:2412.14457}, 2024.

\bibitem{videorag}
X.~Ren et~al.
\newblock Videorag: Retrieval-augmented generation with extreme long-context videos, 2025.

\bibitem{drvideo}
Z.~Ma et~al.
\newblock Drvideo: Document retrieval based long video understanding, 2024.

\bibitem{ctch}
X.~Shen et~al.
\newblock Contrastive transformer cross-modal hashing for video-text retrieval.
\newblock In {\em IJCAI}, 2024.

\bibitem{omagent}
L.~Zhang et~al.
\newblock {O}m{A}gent: A multi-modal agent framework for complex video understanding with task divide-and-conquer.
\newblock In {\em EMNLP}, 2024.

\bibitem{recap}
S.~Ghosh et~al.
\newblock Recap: Retrieval-augmented audio captioning.
\newblock In {\em ICASSP}, 2024.

\bibitem{wavrag}
Y.~Chen et~al.
\newblock Wavrag: Audio-integrated retrieval augmented generation for spoken dialogue models, 2025.

\bibitem{speechrag}
D.~J. Min et~al.
\newblock Speech retrieval-augmented generation without automatic speech recognition.
\newblock In {\em ICASSP}, 2025.

\bibitem{ttarag}
M.~Yang et~al.
\newblock Audiobox tta-rag: Improving zero-shot and few-shot text-to-audio with retrieval-augmented generation, 2024.

\bibitem{mis}
C.~Su et~al.
\newblock Hybrid rag-empowered multi-modal llm for secure data management in internet of medical things: A diffusion-based contract approach.
\newblock {\em IEEE Internet of Things Journal}, 2024.

\bibitem{msier}
Y.~Luo et~al.
\newblock How does the textual information affect the retrieval of multimodal in-context learning?
\newblock {\em arXiv:2404.12866}, 2024.

\bibitem{rule}
P.~Xia et~al.
\newblock Rule: Reliable multimodal rag for factuality in medical vision language models.
\newblock In {\em EMNLP}, 2024.

\bibitem{ramm}
Z.~Yuan et~al.
\newblock Ramm: Retrieval-augmented biomedical visual question answering with multi-modal pre-training.
\newblock In {\em ACMMM}, 2023.

\bibitem{rs}
M.~Mortaheb et~al.
\newblock Re-ranking the context for multimodal retrieval augmented generation.
\newblock {\em arXiv:2501.04695}, 2025.

\bibitem{ragcheck}
M.~Mortaheb and Khojastepour.
\newblock Rag-check: Evaluating multimodal retrieval augmented generation performance.
\newblock {\em arXiv:2501.03995}, 2025.

\bibitem{mrrag}
T.~Zhang et~al.
\newblock mr2ag: Multimodal retrieval-reflection-augmented generation for knowledge-based vqa.
\newblock {\em arXiv:2411.15041}, 2024.

\bibitem{omgqa}
L.~Nan et~al.
\newblock Omg-qa: Building open-domain multi-modal generative question answering systems.
\newblock In {\em EMNLP: Industry Track}, 2024.

\bibitem{ldre}
Z.~Yang et~al.
\newblock Ldre: Llm-based divergent reasoning and ensemble for zero-shot composed image retrieval.
\newblock In {\em International ACM SIGIR Conference on Research and Development in Information Retrieval}, 2024.

\bibitem{unirag}
Q.~Z. Lim et~al.
\newblock Unirag: Unification, retrieval, and generation for multimodal question answering with pre-trained language models.
\newblock {\em IEEE Access}, 2024.

\bibitem{egoinstructor}
J.~Xu et~al.
\newblock Retrieval-augmented egocentric video captioning.
\newblock In {\em CVPR}, 2024.

\bibitem{mmembed}
S.-C. Lin et~al.
\newblock Mm-embed: Universal multimodal retrieval with multimodal llms.
\newblock {\em arXiv:2411.02571}, 2024.

\bibitem{main}
C.-Y. Chang et~al.
\newblock Main-rag: Multi-agent filtering retrieval-augmented generation.
\newblock {\em arXiv:2501.00332}, 2024.

\bibitem{racm3}
M.~Yasunaga et~al.
\newblock Retrieval-augmented multimodal language modeling.
\newblock {\em arXiv:2211.12561}, 2022.

\bibitem{rmr}
C.~Tan et~al.
\newblock Retrieval meets reasoning: Even high-school textbook knowledge benefits multimodal reasoning.
\newblock {\em arXiv:2405.20834}, 2024.

\bibitem{visdom}
M.~Suri et~al.
\newblock Visdom: Multi-document qa with visually rich elements using multimodal retrieval-augmented generation.
\newblock {\em arXiv:2412.10704}, 2024.

\bibitem{samrag}
W.~Zhai.
\newblock Self-adaptive multimodal retrieval-augmented generation.
\newblock {\em arXiv:2410.11321}, 2024.

\bibitem{ragar}
M.~A. Khaliq et~al.
\newblock Ragar, your falsehood radar: Rag-augmented reasoning for political fact-checking using multimodal large language models.
\newblock {\em arXiv:2404.12065}, 2024.

\bibitem{iramig}
X.~Liu et~al.
\newblock Iterative retrieval augmentation for multi-modal knowledge integration and generation.
\newblock {\em Authorea Preprints}, 2024.

\bibitem{ragdriver}
J.~Yuan et~al.
\newblock Rag-driver: Generalisable driving explanations with retrieval-augmented in-context learning in multi-modal large language model.
\newblock {\em arXiv:2402.10828}, 2024.

\bibitem{wikillava}
D.~Caffagni et~al.
\newblock Wiki-llava: Hierarchical retrieval-augmented generation for multimodal llms.
\newblock In {\em CVPR}, 2024.

\bibitem{reflectiva}
F.~Cocchi et~al.
\newblock Augmenting multimodal llms with self-reflective tokens for knowledge-based visual question answering.
\newblock In {\em CVPR}, 2025.

\bibitem{mmedrag}
P.~Xia et~al.
\newblock Mmed-rag: Versatile multimodal rag system for medical vision language models.
\newblock {\em arXiv:2410.13085}, 2024.

\bibitem{ragpt}
J.~Lang et~al.
\newblock Retrieval-augmented dynamic prompt tuning for incomplete multimodal learning.
\newblock {\em arXiv:2501.01120}, 2025.

\bibitem{visionmatters}
Y.~Li et~al.
\newblock Vision matters: Simple visual perturbations can boost multimodal math reasoning.
\newblock {\em arXiv:2506.09736}, 2025.

\bibitem{refocus}
X.~Fu et~al.
\newblock Refocus: Visual editing as a chain of thought for structured image understanding.
\newblock {\em arXiv:2501.05452}, 2025.

\bibitem{skyworkr1v}
Y.~Peng et~al.
\newblock Skywork r1v: Pioneering multimodal reasoning with chain-of-thought.
\newblock {\em arXiv:2504.05599}, 2025.

\bibitem{cmmcot}
G.~Zhang et~al.
\newblock Cmmcot: Enhancing complex multi-image comprehension via multi-modal chain-of-thought and memory augmentation.
\newblock {\em arXiv:2503.05255}, 2025.

\bibitem{mmrlhf}
Y.-F. Zhang et~al.
\newblock Mm-rlhf: The next step forward in multimodal llm alignment.
\newblock {\em arXiv:2502.10391}, 2025.

\bibitem{r1onevision}
Y.~Yang et~al.
\newblock R1-onevision: Advancing generalized multimodal reasoning through cross-modal formalization.
\newblock {\em arXiv:2503.10615}, 2025.

\bibitem{llavacot}
G.~Xu et~al.
\newblock Llava-o1: Let vision language models reason step-by-step.
\newblock {\em arXiv:2411.10440}, 2024.

\bibitem{visualcot}
H.~Shao et~al.
\newblock Visual cot: Advancing multi-modal language models with a comprehensive dataset and benchmark for chain-of-thought reasoning.
\newblock {\em NeurIPS}, 2024.

\bibitem{llavareasoner}
R.~Zhang et~al.
\newblock Improve vision language model chain-of-thought reasoning.
\newblock {\em arXiv:2410.16198}, 2024.

\bibitem{ixc}
Y.~Zang et~al.
\newblock Internlm-xcomposer2. 5-reward: A simple yet effective multi-modal reward model.
\newblock {\em arXiv:2501.12368}, 2025.

\bibitem{insightv}
Y.~Dong et~al.
\newblock Insight-v: Exploring long-chain visual reasoning with multimodal large language models.
\newblock In {\em CVPR}, 2025.

\bibitem{r1vl}
J.~Zhang et~al.
\newblock R1-vl: Learning to reason with multimodal large language models via step-wise group relative policy optimization.
\newblock {\em arXiv:2503.12937}, 2025.

\bibitem{currreft}
H.~Deng et~al.
\newblock Boosting the generalization and reasoning of vision language models with curriculum reinforcement learning.
\newblock {\em arXiv:2503.07065}, 2025.

\bibitem{armcts}
G.~Dong et~al.
\newblock Progressive multimodal reasoning via active retrieval.
\newblock {\em arXiv:2412.14835}, 2024.

\bibitem{aurora}
M.~Bigverdi et~al.
\newblock Perception tokens enhance visual reasoning in multimodal language models.
\newblock In {\em CVPR}, 2025.

\bibitem{vot}
W.~Wu et~al.
\newblock Mind's eye of llms: Visualization-of-thought elicits spatial reasoning in large language models.
\newblock In {\em NeurIPS}, 2024.

\bibitem{dcot}
Z.~Jia et~al.
\newblock Dcot: Dual chain-of-thought prompting for large multimodal models.
\newblock In {\em Asian Conference on Machine Learning (Conference Track)}, 2024.

\bibitem{examplar}
H.~Luo et~al.
\newblock Chain-of-exemplar: enhancing distractor generation for multimodal educational question generation.
\newblock In {\em ACL}, 2024.

\bibitem{dettoolchain}
Y.~Wu et~al.
\newblock Dettoolchain: A new prompting paradigm to unleash detection ability of mllm.
\newblock In {\em ECCV}, 2024.

\bibitem{cantor}
T.~Gao et~al.
\newblock Cantor: Inspiring multimodal chain-of-thought of mllm.
\newblock In {\em ACMMM}, 2024.

\bibitem{got}
Y.~Yao et~al.
\newblock Beyond chain-of-thought, effective graph-of-thought reasoning in language models.
\newblock {\em arXiv:2305.16582}, 2023.

\bibitem{ccot}
C.~Mitra et~al.
\newblock Compositional chain-of-thought prompting for large multimodal models.
\newblock In {\em CVPR}, 2024.

\bibitem{cotdet}
J.~Tang et~al.
\newblock Cotdet: Affordance knowledge prompting for task driven object detection.
\newblock In {\em ICCV}, 2023.

\bibitem{intentqa}
J.~Li et~al.
\newblock Intentqa: Context-aware video intent reasoning.
\newblock In {\em ICCV}, 2023.

\bibitem{antgpt}
Q.~Zhao et~al.
\newblock Antgpt: Can large language models help long-term action anticipation from videos?
\newblock {\em arXiv:2307.16368}, 2023.

\bibitem{vip}
V.~Himakunthala et~al.
\newblock Let's think frame by frame with vip: A video infilling and prediction dataset for evaluating video chain-of-thought.
\newblock {\em arXiv:2305.13903}, 2023.

\bibitem{viot}
H.~Fei et~al.
\newblock Video-of-thought: Step-by-step video reasoning from perception to cognition.
\newblock {\em arXiv:2501.03230}, 2024.

\bibitem{videohal}
Y.~Sun et~al.
\newblock Hallucination mitigation prompts long-term video understanding.
\newblock {\em arXiv:2406.11333}, 2024.

\bibitem{cardiff}
Y.~Tang et~al.
\newblock Cardiff: Video salient object ranking chain of thought reasoning for saliency prediction with diffusion.
\newblock In {\em AAAI}, 2025.

\bibitem{tiprego}
L.~Plini et~al.
\newblock Ti-prego: Chain of thought and in-context learning for online mistake detection in procedural egocentric videos.
\newblock {\em arXiv:2411.02570}, 2024.

\bibitem{exmrd}
R.~Hong et~al.
\newblock Following clues, approaching the truth: Explainable micro-video rumor detection via chain-of-thought reasoning.
\newblock In {\em ACM on Web Conference}, 2025.

\bibitem{cos}
J.~Hu et~al.
\newblock Cos: Chain-of-shot prompting for long video understanding.
\newblock {\em arXiv:2502.06428}, 2025.

\bibitem{videor1}
K.~Feng et~al.
\newblock Video-r1: Reinforcing video reasoning in mllms.
\newblock {\em arXiv:2503.21776}, 2025.

\bibitem{videorft}
Q.~Wang et~al.
\newblock Videorft: Incentivizing video reasoning capability in mllms via reinforced fine-tuning.
\newblock {\em arXiv:2505.12434}, 2025.

\bibitem{unifiedrewardvideo}
Y.~Wang et~al.
\newblock Unified multimodal chain-of-thought reward model through reinforcement fine-tuning.
\newblock {\em arXiv:2505.03318}, 2025.

\bibitem{videomind}
Y.~Liu et~al.
\newblock Videomind: A chain-of-lora agent for long video reasoning.
\newblock {\em arXiv:2503.13444}, 2025.

\bibitem{audiocot}
Z.~Ma et~al.
\newblock Audio-cot: Exploring chain-of-thought reasoning in large audio language model.
\newblock {\em arXiv:2501.07246}, 2025.

\bibitem{cotst}
Y.~Du et~al.
\newblock Cot-st: Enhancing llm-based speech translation with multimodal chain-of-thought.
\newblock {\em arXiv:2409.19510}, 2024.

\bibitem{lpe}
J.~Xie et~al.
\newblock Leveraging chain of thought towards empathetic spoken dialogue without corresponding question-answering data.
\newblock {\em arXiv:2501.10937}, 2025.

\bibitem{spatialsonic}
P.~Sun et~al.
\newblock Both ears wide open: Towards language-driven spatial audio generation.
\newblock {\em arXiv:2410.10676}, 2024.

\bibitem{audioreasoner}
Z.~Xie et~al.
\newblock Audio-reasoner: Improving reasoning capability in large audio language models.
\newblock {\em arXiv:2503.02318}, 2025.

\bibitem{mellow}
S.~Deshmukh et~al.
\newblock Mellow: a small audio language model for reasoning.
\newblock {\em arXiv:2503.08540}, 2025.

\bibitem{grpoaudio}
G.~Li et~al.
\newblock Reinforcement learning outperforms supervised fine-tuning: A case study on audio question answering.
\newblock {\em arXiv:2503.11197}, 2025.

\bibitem{VCD_Ana}
Y.-L. Lee et~al.
\newblock Delve into visual contrastive decoding for hallucination mitigation of large vision-language models.
\newblock {\em arXiv:2412.06775}, 2024.

\bibitem{causality}
P.-H. Huang et~al.
\newblock Who brings the frisbee: Probing hidden hallucination factors in large vision-language model via causality analysis.
\newblock In {\em WACV}, 2025.

\bibitem{ground}
G.~Geigle et~al.
\newblock Does object grounding really reduce hallucination of large vision-language models?
\newblock {\em arXiv:2406.14492}, 2024.

\bibitem{lens}
Z.~Jiang et~al.
\newblock Devils in middle layers of large vision-language models: Interpreting, detecting and mitigating object hallucinations via attention lens.
\newblock In {\em CVPR}, 2025.

\bibitem{fclip}
H.~Oh and W.~Hwang.
\newblock Vision-encoders (already) know what they see: Mitigating object hallucination via simple fine-grained clipscore.
\newblock {\em arXiv:2502.20034}, 2025.

\bibitem{vic}
H.~Zheng et~al.
\newblock Thinking before looking: Improving multimodal llm reasoning via mitigating visual hallucination.
\newblock {\em arXiv:2411.12591}, 2024.

\bibitem{pacu}
M.~Zhao et~al.
\newblock Effectively enhancing vision language large models by prompt augmentation and caption utilization.
\newblock {\em arXiv:2409.14484}, 2024.

\bibitem{mad}
Z.~Lin et~al.
\newblock Interpreting and mitigating hallucination in mllms through multi-agent debate.
\newblock {\em arXiv:2407.20505}, 2024.

\bibitem{fact}
M.~Gao et~al.
\newblock Fact: Teaching mllms with faithful, concise and transferable rationales.
\newblock In {\em ACMMM}, 2024.

\bibitem{pvp}
J.~Kim et~al.
\newblock What if...?: Thinking counterfactual keywords helps to mitigate hallucination in large multi-modal models.
\newblock {\em arXiv:2403.13513}, 2024.

\bibitem{octopus}
W.~Suo et~al.
\newblock Octopus: Alleviating hallucination via dynamic contrastive decoding.
\newblock In {\em CVPR}, 2025.

\bibitem{vhexpansion}
Z.~Liu et~al.
\newblock Automatically generating visual hallucination test cases for multimodal large language models.
\newblock {\em arXiv:2410.11242}, 2024.

\bibitem{dhcp}
Y.~Zhang et~al.
\newblock Dhcp: Detecting hallucinations by cross-modal attention pattern in large vision-language models.
\newblock {\em arXiv:2411.18659}, 2024.

\bibitem{vluncertainty}
R.~Zhang et~al.
\newblock Vl-uncertainty: Detecting hallucination in large vision-language model via uncertainty estimation.
\newblock {\em arXiv:2411.11919}, 2024.

\bibitem{dentist}
Y.~Chang et~al.
\newblock A unified hallucination mitigation framework for large vision-language models.
\newblock {\em arXiv:2409.16494}, 2024.

\bibitem{volcano}
S.~Lee et~al.
\newblock Volcano: mitigating multimodal hallucination through self-feedback guided revision.
\newblock {\em arXiv:2311.07362}, 2023.

\bibitem{vfc}
Y.~Ge et~al.
\newblock Visual fact checker: enabling high-fidelity detailed caption generation.
\newblock In {\em CVPR}, 2024.

\bibitem{hsadpo}
W.~Xiao et~al.
\newblock Detecting and mitigating hallucination in large vision language models via fine-grained ai feedback.
\newblock In {\em AAAI}, 2025.

\bibitem{dftg}
R.~Hu et~al.
\newblock Prescribing the right remedy: Mitigating hallucinations in large vision-language models via targeted instruction tuning.
\newblock {\em Information Sciences}, 2025.

\bibitem{fgaif}
L.~Jing and X.~Du.
\newblock Fgaif: Aligning large vision-language models with fine-grained ai feedback.
\newblock {\em arXiv:2404.05046}, 2024.

\bibitem{seeing}
A.~Deng et~al.
\newblock Seeing is believing: Mitigating hallucination in large vision-language models via clip-guided decoding.
\newblock {\em arXiv:2402.15300}, 2024.

\bibitem{logiccheckgPT}
J.~Wu et~al.
\newblock Logical closed loop: Uncovering object hallucinations in large vision-language models.
\newblock {\em arXiv:2402.11622}, 2024.

\bibitem{eazy}
L.~Che et~al.
\newblock Eazy: Eliminating hallucinations in lvlms by zeroing out hallucinatory image tokens.
\newblock {\em arXiv:2503.07772}, 2025.

\bibitem{efuf}
S.~Xing et~al.
\newblock Efuf: Efficient fine-grained unlearning framework for mitigating hallucinations in multimodal large language models.
\newblock {\em arXiv:2402.09801}, 2024.

\bibitem{hadpo}
Z.~Zhao et~al.
\newblock Beyond hallucinations: Enhancing lvlms through hallucination-aware direct preference optimization.
\newblock {\em arXiv:2311.16839}, 2023.

\bibitem{hallucidoctor}
Q.~Yu et~al.
\newblock Hallucidoctor: Mitigating hallucinatory toxicity in visual instruction data.
\newblock In {\em CVPR}, 2024.

\bibitem{buhr}
S.~Wu et~al.
\newblock Combating multimodal llm hallucination via bottom-up holistic reasoning.
\newblock In {\em AAAI}, 2025.

\bibitem{catch}
Z.~Kan et~al.
\newblock Catch: Complementary adaptive token-level contrastive decoding to mitigate hallucinations in lvlms.
\newblock {\em arXiv:2411.12713}, 2024.

\bibitem{vti}
S.~Liu et~al.
\newblock Reducing hallucinations in vision-language models via latent space steering.
\newblock {\em arXiv:2410.15778}, 2024.

\bibitem{ara}
X.~Qu et~al.
\newblock Alleviating hallucination in large vision-language models with active retrieval augmentation.
\newblock {\em ACM Transactions on Multimedia Computing, Communications and Applications}, 2024.

\bibitem{bacon}
Z.~Yang et~al.
\newblock Bacon: Supercharge your vlm with bag-of-concept graph to mitigate hallucinations.
\newblock {\em arXiv:2407.03314}, 2024.

\bibitem{pelican}
P.~Sahu et~al.
\newblock Pelican: Correcting hallucination in vision-llms via claim decomposition and program of thought verification.
\newblock {\em arXiv:2407.02352}, 2024.

\bibitem{agla}
W.~An et~al.
\newblock Agla: Mitigating object hallucinations in large vision-language models with assembly of global and local attention.
\newblock {\em arXiv:2406.12718}, 2024.

\bibitem{noiseboost}
K.~Wu et~al.
\newblock Noiseboost: Alleviating hallucination with noise perturbation for multimodal large language models.
\newblock {\em arXiv:2405.20081}, 2024.

\bibitem{halc}
Z.~Chen et~al.
\newblock Halc: Object hallucination reduction via adaptive focal-contrast decoding.
\newblock {\em arXiv:2403.00425}, 2024.

\bibitem{sumgd}
K.~Min et~al.
\newblock Mitigating hallucinations in large vision-language models via summary-guided decoding.
\newblock {\em arXiv:2410.13321}, 2024.

\bibitem{woodpecker}
S.~Yin et~al.
\newblock Woodpecker: Hallucination correction for multimodal large language models.
\newblock {\em Science China Information Sciences}, 2024.

\bibitem{eagle}
A.~Villa et~al.
\newblock Eagle: Enhanced visual grounding minimizes hallucinations in instructional multimodal models.
\newblock {\em arXiv:2501.02699}, 2025.

\bibitem{llavarlhf}
Z.~Sun et~al.
\newblock Aligning large multimodal models with factually augmented rlhf.
\newblock {\em arXiv:2309.14525}, 2023.

\bibitem{hdpo}
Y.~Fu et~al.
\newblock Mitigating hallucination in multimodal large language model via hallucination-targeted direct preference optimization.
\newblock {\em arXiv:2411.10436}, 2024.

\bibitem{vdpo}
Y.~Xie et~al.
\newblock V-dpo: Mitigating hallucination in large vision language models via vision-guided direct preference optimization.
\newblock {\em arXiv:2411.02712}, 2024.

\bibitem{helpd}
F.~Yuan et~al.
\newblock Helpd: Mitigating hallucination of lvlms by hierarchical feedback learning with vision-enhanced penalty decoding.
\newblock {\em arXiv:2409.20429}, 2024.

\bibitem{clipdpo}
Y.~Ouali et~al.
\newblock Clip-dpo: Vision-language models as a source of preference for fixing hallucinations in lvlms.
\newblock In {\em ECCV}, 2024.

\bibitem{reverie}
J.~Zhang et~al.
\newblock Reflective instruction tuning: Mitigating hallucinations in large vision-language models.
\newblock In {\em ECCV}, 2024.

\bibitem{comt}
Y.~Jiang et~al.
\newblock Comt: Chain-of-medical-thought reduces hallucination in medical report generation.
\newblock {\em arXiv:2406.11451}, 2024.

\bibitem{rlaifv}
T.~Yu et~al.
\newblock Rlaif-v: Open-source ai feedback leads to super gpt-4v trustworthiness.
\newblock In {\em CVPR}, 2025.

\bibitem{povid}
Y.~Zhou et~al.
\newblock Aligning modalities in vision large language models via preference fine-tuning.
\newblock {\em arXiv:2402.11411}, 2024.

\bibitem{hacl}
C.~Jiang et~al.
\newblock Hallucination augmented contrastive learning for multimodal large language model.
\newblock In {\em CVPR}, 2024.

\bibitem{attacklvlm}
Y.~Zhao et~al.
\newblock On evaluating adversarial robustness of large vision-language models.
\newblock {\em NeurIPS}, 2023.

\bibitem{pm}
L.~Bailey et~al.
\newblock Image hijacks: Adversarial images can control generative models at runtime.
\newblock {\em arXiv:2309.00236}, 2023.

\bibitem{attackbard}
Y.~Dong et~al.
\newblock How robust is google's bard to adversarial image attacks?
\newblock {\em arXiv:2309.11751}, 2023.

\bibitem{figstep}
Y.~Gong et~al.
\newblock Figstep: Jailbreaking large vision-language models via typographic visual prompts.
\newblock In {\em AAAI}, 2025.

\bibitem{instructta}
X.~Wang et~al.
\newblock Instructta: Instruction-tuned targeted attack for large vision-language models.
\newblock {\em arXiv:2312.01886}, 2023.

\bibitem{verboseimages}
K.~Gao et~al.
\newblock Inducing high energy-latency of large vision-language models with verbose images.
\newblock {\em arXiv:2401.11170}, 2024.

\bibitem{sgta}
M.~Qraitem et~al.
\newblock Vision-llms can fool themselves with self-generated typographic attacks.
\newblock {\em arXiv:2402.00626}, 2024.

\bibitem{jailbreaking}
Z.~Niu et~al.
\newblock Jailbreaking attack against multimodal large language model.
\newblock {\em arXiv:2402.02309}, 2024.

\bibitem{visco}
Z.~Miao et~al.
\newblock Visual contextual attack: Jailbreaking mllms with image-driven context injection.
\newblock {\em arXiv:2507.02844}, 2025.

\bibitem{shadowcast}
Y.~Xu et~al.
\newblock Shadowcast: Stealthy data poisoning attacks against vision-language models.
\newblock {\em arXiv:2402.06659}, 2024.

\bibitem{badvlmdriver}
Z.~Ni et~al.
\newblock Physical backdoor attack can jeopardize driving with vision-large-language models.
\newblock {\em arXiv:2404.12916}, 2024.

\bibitem{vrp}
S.~Ma et~al.
\newblock Visual-roleplay: Universal jailbreak attack on multimodal large language models via role-playing image character.
\newblock {\em arXiv:2405.20773}, 2024.

\bibitem{bmap}
Z.~Ying et~al.
\newblock Jailbreak vision language models via bi-modal adversarial prompt.
\newblock {\em arXiv:2406.04031}, 2024.

\bibitem{ideator}
R.~Wang et~al.
\newblock Ideator: Jailbreaking large vision-language models using themselves.
\newblock {\em arXiv:2411.00827}, 2024.

\bibitem{himrd}
M.~Teng et~al.
\newblock Heuristic-induced multimodal risk distribution jailbreak attack for multimodal large language models.
\newblock {\em arXiv:2412.05934}, 2024.

\bibitem{badtoken}
Z.~Yuan et~al.
\newblock Badtoken: Token-level backdoor attacks to multi-modal large language models.
\newblock In {\em CVPR}, 2025.

\bibitem{dress}
Y.~Chen et~al.
\newblock Dress: Instructing large vision-language models to align and interact with humans via natural language feedback.
\newblock In {\em CVPR}, 2024.

\bibitem{mllmproctor}
R.~Pi et~al.
\newblock Mllm-protector: Ensuring mllm's safety without hurting performance.
\newblock {\em arXiv:2401.02906}, 2024.

\bibitem{adashield}
Y.~Wang et~al.
\newblock Adashield: Safeguarding multimodal large language models from structure-based attack via adaptive shield prompting.
\newblock In {\em ECCV}, 2024.

\bibitem{safevlm}
Z.~Liu et~al.
\newblock Safety alignment for vision language models.
\newblock {\em arXiv:2405.13581}, 2024.

\bibitem{mirrorcheck}
S.~Fares et~al.
\newblock Mirrorcheck: Efficient adversarial defense for vision-language models.
\newblock {\em arXiv:2406.09250}, 2024.

\bibitem{cider1}
Y.~Xu et~al.
\newblock Cross-modality information check for detecting jailbreaking in multimodal large language models.
\newblock {\em arXiv:2407.21659}, 2024.

\bibitem{pip1}
Y.~Zhang et~al.
\newblock Pip: Detecting adversarial examples in large vision-language models via attention patterns of irrelevant probe questions.
\newblock In {\em ACMMM}, 2024.

\bibitem{vlmguard}
X.~Du et~al.
\newblock Vlmguard: Defending vlms against malicious prompts via unlabeled data.
\newblock {\em arXiv:2410.00296}, 2024.

\bibitem{eta}
Y.~Ding et~al.
\newblock Eta: Evaluating then aligning safety of vision language models at inference time.
\newblock {\em arXiv:2410.06625}, 2024.

\bibitem{bluesuffix}
Y.~Zhao et~al.
\newblock Bluesuffix: Reinforced blue teaming for vision-language models against jailbreak attacks.
\newblock {\em arXiv:2410.20971}, 2024.

\bibitem{immune}
S.~S. Ghosal et~al.
\newblock Immune: Improving safety against jailbreaks in multi-modal llms via inference-time alignment.
\newblock In {\em CVPR}, 2025.

\bibitem{adpo}
F.~Weng et~al.
\newblock Adversary-aware dpo: Enhancing safety alignment in vision language models via adversarial training.
\newblock {\em arXiv:2502.11455}, 2025.

\bibitem{sea}
W.~Lu et~al.
\newblock Sea: Low-resource safety alignment for multimodal large language models via synthetic embeddings.
\newblock {\em arXiv:2502.12562}, 2025.

\bibitem{vlmguardr1}
M.~Chen et~al.
\newblock Vlmguard-r1: Proactive safety alignment for vlms via reasoning-driven prompt optimization.
\newblock {\em arXiv:2504.12661}, 2025.

\bibitem{holisafe}
Y.~Lee et~al.
\newblock Holisafe: Holistic safety benchmarking and modeling with safety meta token for vision-language model.
\newblock {\em arXiv:2506.04704}, 2025.

\bibitem{craft}
L.~Yuan et~al.
\newblock Craft: Customizing llms by creating and retrieving from specialized toolsets.
\newblock {\em arXiv:2309.17428}, 2023.

\bibitem{vipergpt}
D.~Sur{\'\i}s et~al.
\newblock Vipergpt: Visual inference via python execution for reasoning.
\newblock In {\em ICCV}, 2023.

\bibitem{clova}
Z.~Gao et~al.
\newblock Clova: A closed-loop visual assistant with tool usage and update.
\newblock In {\em CVPR}, 2024.

\bibitem{hugginggpt}
Y.~Shen et~al.
\newblock Hugginggpt: Solving ai tasks with chatgpt and its friends in hugging face.
\newblock {\em NeurIPS}, 2023.

\bibitem{chameleon}
P.~Lu et~al.
\newblock Chameleon: Plug-and-play compositional reasoning with large language models.
\newblock {\em NeurIPS}, 2023.

\bibitem{visualchatgpt}
C.~Wu et~al.
\newblock Visual chatgpt: Talking, drawing and editing with visual foundation models.
\newblock {\em arXiv:2303.04671}, 2023.

\bibitem{assistgpt}
D.~Gao et~al.
\newblock Assistgpt: A general multi-modal assistant that can plan, execute, inspect, and learn.
\newblock {\em arXiv:2306.08640}, 2023.

\bibitem{grid}
S.~Vemprala et~al.
\newblock Grid: A platform for general robot intelligence development.
\newblock {\em arXiv:2310.00887}, 2023.

\bibitem{drive}
D.~Fu et~al.
\newblock Drive like a human: Rethinking autonomous driving with large language models.
\newblock In {\em WACVW}, 2024.

\bibitem{assistgui}
D.~Gao et~al.
\newblock Assistgui: Task-oriented desktop graphical user interface automation.
\newblock {\em arXiv:2312.13108}, 2023.

\bibitem{musicagent}
D.~Yu et~al.
\newblock Musicagent: An ai agent for music understanding and generation with large language models.
\newblock {\em arXiv:2310.11954}, 2023.

\bibitem{audiogpt}
R.~Huang et~al.
\newblock Audiogpt: Understanding and generating speech, music, sound, and talking head.
\newblock In {\em AAAI}, 2024.

\bibitem{droidbot}
H.~Wen et~al.
\newblock Droidbot-gpt: Gpt-powered ui automation for android.
\newblock {\em arXiv:2304.07061}, 2023.

\bibitem{deps}
Z.~Wang et~al.
\newblock Describe, explain, plan and select: Interactive planning with large language models enables open-world multi-task agents.
\newblock {\em arXiv:2302.01560}, 2023.

\bibitem{cradle}
W.~Tan et~al.
\newblock Cradle: Empowering foundation agents towards general computer control.
\newblock {\em arXiv:2403.03186}, 2024.

\bibitem{mobileagent}
J.~Wang et~al.
\newblock Mobile-agent: Autonomous multi-modal mobile device agent with visual perception.
\newblock {\em arXiv:2401.16158}, 2024.

\bibitem{seeact}
B.~Zheng et~al.
\newblock Gpt-4v (ision) is a generalist web agent, if grounded.
\newblock {\em arXiv:2401.01614}, 2024.

\bibitem{doraemongpt}
Z.~Yang et~al.
\newblock Doraemongpt: Toward understanding dynamic scenes with large language models (exemplified as a video agent).
\newblock {\em arXiv:2401.08392}, 2024.

\bibitem{chatvideo}
J.~Wang et~al.
\newblock Chatvideo: A tracklet-centric multimodal and versatile video understanding system.
\newblock {\em arXiv:2304.14407}, 2023.

\bibitem{appagent}
C.~Zhang et~al.
\newblock Appagent: Multimodal agents as smartphone users.
\newblock In {\em CHI Conference on Human Factors in Computing Systems}, 2025.

\bibitem{mmnavigator}
A.~Yan et~al.
\newblock Gpt-4v in wonderland: Large multimodal models for zero-shot smartphone gui navigation.
\newblock {\em arXiv:2311.07562}, 2023.

\bibitem{wavjourney}
X.~Liu et~al.
\newblock Wavjourney: Compositional audio creation with large language models.
\newblock {\em TASLP}, 2025.

\bibitem{llavainteractive}
W.-G. Chen et~al.
\newblock Llava-interactive: An all-in-one demo for image chat, segmentation, generation and editing.
\newblock {\em arXiv:2311.00571}, 2023.

\bibitem{jarvis1}
Z.~Wang et~al.
\newblock Jarvis-1: Open-world multi-task agents with memory-augmented multimodal language models.
\newblock {\em TPAMI}, 2024.

\bibitem{steve}
Z.~Zhao et~al.
\newblock See and think: Embodied agent in virtual environment.
\newblock In {\em ECCV}, 2024.

\bibitem{emma}
Y.~Yang et~al.
\newblock Embodied multi-modal agent trained by an llm from a parallel textworld.
\newblock In {\em CVPR}, 2024.

\bibitem{mllmtool}
C.~Wang et~al.
\newblock Mllm-tool: A multimodal large language model for tool agent learning.
\newblock In {\em WACV}, 2025.

\bibitem{llavaplus}
S.~Liu et~al.
\newblock Llava-plus: Learning to use tools for creating multimodal agents.
\newblock In {\em ECCV}, 2024.

\bibitem{gpt4tools}
R.~Yang et~al.
\newblock Gpt4tools: Teaching large language model to use tools via self-instruction.
\newblock {\em NeurIPS}, 2023.

\bibitem{webwise}
H.~Tao et~al.
\newblock Webwise: Web interface control and sequential exploration with large language models.
\newblock {\em arXiv:2310.16042}, 2023.

\bibitem{autogui}
Z.~Zhang and A.~Zhang.
\newblock You only look at screens: Multimodal chain-of-action agents.
\newblock {\em arXiv:2309.11436}, 2023.

\bibitem{zheng2024training}
M.~Zheng et~al.
\newblock Training-free video temporal grounding using large-scale pre-trained models.
\newblock In {\em ECCV}, 2024.

\bibitem{qu2024chatvtg}
M.~Qu et~al.
\newblock Chatvtg: Video temporal grounding via chat with video dialogue large language models.
\newblock In {\em CVPRW}, 2024.

\bibitem{qian2024momentor}
L.~Qian et~al.
\newblock Momentor: Advancing video large language model with fine-grained temporal reasoning.
\newblock In {\em ICML}, 2024.

\bibitem{guo2025vtg}
Y.~Guo et~al.
\newblock Vtg-llm: Integrating timestamp knowledge into video llms for enhanced video temporal grounding.
\newblock In {\em AAAI}, 2025.

\bibitem{liu2024bench}
Y.~Liu et~al.
\newblock Et bench: Towards open-ended event-level video-language understanding.
\newblock In {\em NeurIPS}, 2024.

\bibitem{guo2024trace}
Y.~Guo et~al.
\newblock Trace: Temporal grounding video llm via causal event modeling.
\newblock In {\em ICLR}, 2025.

\bibitem{wang2025time}
Y.~Wang et~al.
\newblock Time-r1: Post-training large vision language model for temporal video grounding.
\newblock {\em arXiv:2503.13377}, 2025.

\bibitem{li2025videochat}
X.~Li et~al.
\newblock Videochat-r1: Enhancing spatio-temporal perception via reinforcement fine-tuning.
\newblock {\em arXiv:2504.06958}, 2025.

\bibitem{luo2025museg}
F.~Luo et~al.
\newblock Museg: Reinforcing video temporal understanding via timestamp-aware multi-segment grounding.
\newblock {\em arXiv:2505.20715}, 2025.

\bibitem{chen2025versavid}
X.~Chen et~al.
\newblock Versavid-r1: A versatile video understanding and reasoning model from question answering to captioning tasks.
\newblock {\em arXiv:2506.09079}, 2025.

\bibitem{meng2025videocap}
D.~Meng et~al.
\newblock Videocap-r1: Enhancing mllms for video captioning via structured thinking.
\newblock {\em arXiv:2506.01725}, 2025.

\bibitem{lin2025perceive}
W.~Lin et~al.
\newblock Perceive anything: Recognize, explain, caption, and segment anything in images and videos.
\newblock {\em arXiv:2506.05302}, 2025.

\bibitem{min2024morevqa}
J.~Min et~al.
\newblock Morevqa: Exploring modular reasoning models for video question answering.
\newblock In {\em CVPR}, 2024.

\bibitem{park2025deepvideo}
J.~Park et~al.
\newblock Deepvideo-r1: Video reinforcement fine-tuning via difficulty-aware regressive grpo.
\newblock {\em arXiv:2506.07464}, 2025.

\bibitem{maaz2024videogpt+}
M.~Maaz et~al.
\newblock Videogpt+: Integrating image and video encoders for enhanced video understanding.
\newblock {\em arXiv:2406.09418}, 2024.

\bibitem{shang2024traveler}
C.~Shang et~al.
\newblock Traveler: A modular multi-lmm agent framework for video question-answering.
\newblock {\em arXiv:2404.01476}, 2024.

\bibitem{videoagent}
Y.~Fan et~al.
\newblock Videoagent: A memory-augmented multimodal agent for video understanding.
\newblock In {\em ECCV}, 2024.

\bibitem{dang2025mupa}
J.~Dang et~al.
\newblock Mupa: Towards multi-path agentic reasoning for grounded video question answering.
\newblock {\em arXiv:2506.18071}, 2025.

\bibitem{kahatapitiya2024language}
K.~Kahatapitiya et~al.
\newblock Language repository for long video understanding.
\newblock {\em arXiv:2403.14622}, 2024.

\bibitem{azad2025hierarq}
S.~Azad et~al.
\newblock Hierarq: Task-aware hierarchical q-former for enhanced video understanding.
\newblock In {\em CVPR}, 2025.

\bibitem{ranasinghe2024understanding}
K.~Ranasinghe et~al.
\newblock Understanding long videos with multimodal language models.
\newblock {\em arXiv:2403.16998}, 2024.

\bibitem{zhang2023simple}
C.~Zhang et~al.
\newblock A simple llm framework for long-range video question-answering.
\newblock {\em arXiv:2312.17235}, 2023.

\bibitem{park2024too}
J.~Park et~al.
\newblock Too many frames, not all useful: Efficient strategies for long-form video qa.
\newblock {\em arXiv:2406.09396}, 2024.

\bibitem{zhang2024omagent}
L.~Zhang et~al.
\newblock Omagent: A multi-modal agent framework for complex video understanding with task divide-and-conquer.
\newblock {\em arXiv:2406.16620}, 2024.

\bibitem{mvot}
C.~Li et~al.
\newblock Imagine while reasoning in space: Multimodal visualization-of-thought.
\newblock {\em arXiv:2501.07542}, 2025.

\bibitem{hallucinationsurvey}
Z.~Bai et~al.
\newblock Hallucination of multimodal large language models: A survey.
\newblock {\em arXiv:2404.18930}, 2024.

\bibitem{xing2024mitigating}
Y.~Xing et~al.
\newblock Mitigating object hallucination via concentric causal attention.
\newblock In {\em NeurIPS}, 2024.

\bibitem{zhang2024seeing}
X.~Zhang et~al.
\newblock Seeing clearly by layer two: Enhancing attention heads to alleviate hallucination in lvlms.
\newblock {\em arXiv:2411.09968}, 2024.

\bibitem{ho2020denoising}
J.~Ho et~al.
\newblock Denoising diffusion probabilistic models.
\newblock {\em NeurIPS}, 2020.

\bibitem{brooks2023instructpix2pix}
T.~Brooks et~al.
\newblock Instructpix2pix: Learning to follow image editing instructions.
\newblock In {\em CVPR}, 2023.

\bibitem{yu2023inpaint}
T.~Yu et~al.
\newblock Inpaint anything: Segment anything meets image inpainting.
\newblock {\em arXiv:2304.06790}, 2023.

\bibitem{sam}
A.~Kirillov et~al.
\newblock Segment anything.
\newblock In {\em ICCV}, 2023.

\bibitem{galore}
J.~Zhao et~al.
\newblock Galore: Memory-efficient llm training by gradient low-rank projection.
\newblock {\em arXiv:2403.03507}, 2024.

\bibitem{dpo}
R.~Rafailov et~al.
\newblock Direct preference optimization: Your language model is secretly a reward model.
\newblock {\em NeurIPS}, 2023.

\bibitem{ddpm}
P.~Dhariwal and A.~Nichol.
\newblock Diffusion models beat gans on image synthesis.
\newblock {\em NeurIPS}, 2021.

\bibitem{owlv2}
M.~Minderer et~al.
\newblock Scaling open-vocabulary object detection.
\newblock {\em NeurIPS}, 36:72983--73007, 2023.

\bibitem{cd}
X.~L. Li et~al.
\newblock Contrastive decoding: Open-ended text generation as optimization.
\newblock In {\em ACL}, 2023.

\bibitem{t5}
C.~Raffel et~al.
\newblock Exploring the limits of transfer learning with a unified text-to-text transformer.
\newblock {\em Journal of machine learning research}, 2020.

\bibitem{internvl25}
Z.~Chen et~al.
\newblock Expanding performance boundaries of open-source multimodal models with model, data, and test-time scaling.
\newblock {\em arXiv:2412.05271}, 2024.

\bibitem{glip}
L.~H. Li et~al.
\newblock Grounded language-image pre-training.
\newblock In {\em CVPR}, 2022.

\bibitem{xvlm}
Y.~Zeng et~al.
\newblock Multi-grained vision language pre-training: Aligning texts with visual concepts.
\newblock {\em arXiv:2111.08276}, 2021.

\bibitem{gligen}
Y.~Li et~al.
\newblock Gligen: Open-set grounded text-to-image generation.
\newblock In {\em CVPR}, 2023.

\bibitem{trihe}
J.~Wu et~al.
\newblock Unified triplet-level hallucination evaluation for large vision-language models.
\newblock {\em arXiv:2410.23114}, 2024.

\bibitem{amber}
J.~Wang et~al.
\newblock Amber: An llm-free multi-dimensional benchmark for mllms hallucination evaluation.
\newblock {\em arXiv:2311.07397}, 2023.

\bibitem{nltk}
E.~Loper and S.~Bird.
\newblock Nltk: The natural language toolkit.
\newblock {\em cs/0205028}, 2002.

\bibitem{wordnet}
G.~A. Miller.
\newblock Wordnet: a lexical database for english.
\newblock {\em Communications of the ACM}, 1995.

\bibitem{honnibal2020spacy}
M.~Honnibal et~al.
\newblock spacy: Industrial-strength natural language processing in python.
\newblock {\em Zenodo, Honolulu, HI, USA}, 2020.

\bibitem{jing2023faithscore}
L.~Jing et~al.
\newblock Faithscore: Fine-grained evaluations of hallucinations in large vision-language models.
\newblock {\em arXiv:2311.01477}, 2023.

\bibitem{ssgp}
S.~Schuster et~al.
\newblock Generating semantically precise scene graphs from textual descriptions for improved image retrieval.
\newblock In {\em Proceedings of the fourth workshop on vision and language}, 2015.

\bibitem{pcfg}
D.~Klein and C.~D. Manning.
\newblock Accurate unlexicalized parsing.
\newblock In {\em ACL}, 2003.

\bibitem{spice}
P.~Anderson et~al.
\newblock Spice: Semantic propositional image caption evaluation.
\newblock In {\em ECCV}, 2016.

\bibitem{ragchecker}
D.~Ru et~al.
\newblock Ragchecker: A fine-grained framework for diagnosing retrieval-augmented generation.
\newblock {\em NeurIPS}, 2024.

\bibitem{haloquest}
Z.~Wang et~al.
\newblock Haloquest: A visual hallucination dataset for advancing multimodal reasoning.
\newblock In {\em ECCV}, 2024.

\bibitem{valor}
H.~Qiu et~al.
\newblock Valor-eval: Holistic coverage and faithfulness evaluation of large vision-language models.
\newblock {\em arXiv:2404.13874}, 2024.

\bibitem{ragas}
S.~Es et~al.
\newblock Ragas: Automated evaluation of retrieval augmented generation.
\newblock In {\em Conference of the European Chapter of the Association for Computational Linguistics: System Demonstrations}, 2024.

\bibitem{fiha}
B.~Yan et~al.
\newblock Fiha: Autonomous hallucination evaluation in vision-language models with davidson scene graphs.
\newblock {\em arXiv:2409.13612}, 2024.

\bibitem{pfram}
Y.~Wang et~al.
\newblock Understanding multimodal hallucination with parameter-free representation alignment.
\newblock {\em arXiv:2409.01151}, 2024.

\bibitem{reefknot}
K.~Zheng et~al.
\newblock Reefknot: A comprehensive benchmark for relation hallucination evaluation, analysis and mitigation in multimodal large language models.
\newblock {\em arXiv:2408.09429}, 2024.

\bibitem{dbd}
M.~Feng et~al.
\newblock Do more details always introduce more hallucinations in lvlm-based image captioning?
\newblock {\em arXiv:2406.12663}, 2024.

\bibitem{medihalldetector}
J.~Chen et~al.
\newblock Detecting and evaluating medical hallucinations in large vision language models.
\newblock {\em arXiv:2406.10185}, 2024.

\bibitem{metatoken}
L.~Fieback et~al.
\newblock Metatoken: Detecting hallucination in image descriptions by meta classification.
\newblock {\em arXiv:2405.19186}, 2024.

\bibitem{avhbench}
K.~Sung-Bin et~al.
\newblock Avhbench: A cross-modal hallucination benchmark for audio-visual large language models.
\newblock {\em arXiv:2410.18325}, 2024.

\bibitem{vcd}
S.~Leng et~al.
\newblock Mitigating object hallucinations in large vision-language models through visual contrastive decoding.
\newblock {\em arXiv:2311.16922}, 2023.

\bibitem{hallusionbench}
T.~Guan et~al.
\newblock Hallusionbench: an advanced diagnostic suite for entangled language hallucination and visual illusion in large vision-language models.
\newblock In {\em CVPR}, 2024.

\bibitem{vlmevalkit}
H.~Duan et~al.
\newblock Vlmevalkit: An open-source toolkit for evaluating large multi-modality models.
\newblock In {\em ACMMM}, 2024.

\bibitem{wildvision}
Y.~Lu et~al.
\newblock Wildvision: Evaluating vision-language models in the wild with human preferences.
\newblock {\em NeurIPS}, 2024.

\bibitem{lmmseval}
K.~Zhang et~al.
\newblock Lmms-eval: Reality check on the evaluation of large multimodal models, 2024.

\bibitem{multimedeval}
C.~Royer et~al.
\newblock Multimedeval: A benchmark and a toolkit for evaluating medical vision-language models, 2024.

\bibitem{agentstudio}
L.~Zheng et~al.
\newblock Agentstudio: A toolkit for building general virtual agents.
\newblock {\em arXiv:2403.17918}, 2024.

\bibitem{videobench}
M.~Ning et~al.
\newblock Video-bench: A comprehensive benchmark and toolkit for evaluating video-based large language models.
\newblock {\em arXiv:2311.16103}, 2023.

\bibitem{llavanext}
B.~Li et~al.
\newblock Llava-next: Stronger llms supercharge multimodal capabilities in the wild.
\newblock {\em arXiv preprint}, 2024.

\bibitem{ren2024timechat}
S.~Ren et~al.
\newblock Timechat: A time-sensitive multimodal large language model for long video understanding.
\newblock In {\em CVPR}, 2024.

\bibitem{zhang2023video}
H.~Zhang et~al.
\newblock Video-llama: An instruction-tuned audio-visual language model for video understanding.
\newblock In {\em EMNLP Demo}, 2023.

\bibitem{lin2023video}
B.~Lin et~al.
\newblock Video-llava: Learning united visual representation by alignment before projection.
\newblock In {\em EMNLP}, 2024.

\bibitem{gao2017tall}
J.~Gao et~al.
\newblock Tall: Temporal activity localization via language query.
\newblock In {\em ICCV}, 2017.

\bibitem{sentencebert}
N.~Reimers and I.~Gurevych.
\newblock Sentence-bert: Sentence embeddings using siamese bert-networks.
\newblock In {\em EMNLP}, 2019.

\bibitem{krishna2017dense}
R.~Krishna et~al.
\newblock Dense-captioning events in videos.
\newblock In {\em ICCV}, 2017.

\bibitem{li2024mvbench}
K.~Li et~al.
\newblock Mvbench: A comprehensive multi-modal video understanding benchmark.
\newblock In {\em CVPR}, 2024.

\bibitem{fu2024video}
C.~Fu et~al.
\newblock Video-mme: The first-ever comprehensive evaluation benchmark of multi-modal llms in video analysis.
\newblock {\em arXiv:2405.21075}, 2024.

\bibitem{minderer2022simple}
M.~Minderer et~al.
\newblock Simple open-vocabulary object detection.
\newblock In {\em ECCV}, 2022.

\bibitem{zhao2023learning}
Y.~Zhao et~al.
\newblock Learning video representations from large language models.
\newblock In {\em CVPR}, 2023.

\end{thebibliography}
}

\end{document}